\documentclass{article}

\usepackage{microtype}
\usepackage{graphicx}
\usepackage{subcaption}
\usepackage{booktabs} 
\usepackage{bm}
\usepackage{xcolor}
\usepackage{multirow}

\usepackage{hyperref}

\usepackage[preprint]{icml2026}

\usepackage{amsmath}
\usepackage{amssymb}
\usepackage{mathtools}
\usepackage{amsthm}

\usepackage[capitalize,noabbrev]{cleveref}

\theoremstyle{plain}

\theoremstyle{definition}

\theoremstyle{remark}

\usepackage[textsize=tiny]{todonotes}

\icmltitlerunning{Progressive Refinement Regulation for Accelerating Diffusion Language Model Decoding}
\usepackage[most]{tcolorbox}

\definecolor{qbgb}{RGB}{235,245,255}
\definecolor{qbfr}{RGB}{60,120,180}
\definecolor{abgb}{RGB}{235,250,240}
\definecolor{abfr}{RGB}{60,140,90}
\definecolor{rbgb}{RGB}{255,250,230}
\definecolor{rbfr}{RGB}{190,120,40}
\definecolor{basegb}{RGB}{255,235,238} 
\definecolor{basefr}{RGB}{198,40,40}   
\definecolor{mbgb}{RGB}{245,245,245}
\definecolor{mbfr}{RGB}{120,120,120}

\tcbset{
  icmlboxstyle/.style={
    enhanced,
    breakable,
    sharp corners,
    boxrule=0.8pt,
    left=6pt,right=6pt,top=6pt,bottom=6pt,
    before skip=6pt, after skip=6pt,
    coltitle=black,
    fonttitle=\bfseries,
    attach title to upper,
    title={#1},
  }
}

\newtcolorbox{questionbox}[1]{icmlboxstyle={#1}, colframe=qbfr, colback=qbgb}
\newtcolorbox{answerbox}{icmlboxstyle={Ground-truth answer}, colframe=abfr, colback=abgb}
\newtcolorbox{baselinebox}[1]{icmlboxstyle={#1}, colframe=rbfr, colback=rbgb}
\newtcolorbox{reasoningbox}[1]{icmlboxstyle={#1}, colframe=basefr, colback=basegb}
\newtcolorbox{metricsbox}{icmlboxstyle={Metrics}, colframe=mbfr, colback=mbgb}

\newcommand{\stepword}[2]{\ensuremath{\overset{\text{\tiny\color{blue}#2}}{\text{#1}}}}
\newcommand{\finalanswer}[1]{\ensuremath{\boxed{#1}}}

\begin{document}

\twocolumn[
  \icmltitle{Progressive Refinement Regulation for Accelerating Diffusion Language Model Decoding}



  \icmlsetsymbol{equal}{*}

  \begin{icmlauthorlist}
    \icmlauthor{Lipeng Wan}{equal,xjtu}
    \icmlauthor{Jianhui Gu}{equal,xjtu}
    \icmlauthor{Junjie Ma}{xjtu}
    \icmlauthor{Jianguo Huang}{ntu}
    \icmlauthor{Shiguang Sun}{xjtu}
    \icmlauthor{Siyuan Li}{hit}
    \icmlauthor{Xuguang Lan}{xjtu}
\end{icmlauthorlist}

\icmlaffiliation{xjtu}{
Department of Artificial Intelligence,
Xi'an Jiaotong University,
Xi'an, Shaanxi, China
}

\icmlaffiliation{ntu}{
College of Computing and Data Science,
Nanyang Technological University,
Singapore
}

\icmlaffiliation{hit}{
School of Computer Science and Technology,
Harbin Institute of Technology,
Harbin, China
}

\icmlcorrespondingauthor{Xuguang Lan}{xglan@xjtu.edu.cn}

  \icmlkeywords{Diffusion Language Models}

  \vskip 0.3in
]



\printAffiliationsAndNotice{}  

\begin{abstract}
Diffusion language models generate text through iterative denoising under a uniform refinement rule applied to all tokens. 
However, tokens stabilize at different rates in practice, leading to substantial redundant refinement and motivating refinement control over the denoising process. 
Existing approaches typically assess refinement necessity from instantaneous, step-level signals under a fixed decoding process. 
In contrast, whether a token has converged is defined by how its prediction changes along its future refinement trajectory.
Moreover, changing the refinement rule reshapes future refinement trajectories, which in turn determine how refinement rules should be formulated, making refinement control inherently dynamic.
We propose \emph{Progressive Refinement Regulation} (PRR), a progressive, trajectory-grounded refinement control framework that derives a token-level notion of empirical convergence progress from full decoding rollouts. 
Based on this signal, PRR learns a lightweight token-wise controller to regulate refinement via temperature-based distribution shaping under a progressive self-evolving training scheme. 
Experiments show that PRR substantially accelerates diffusion language model decoding while preserving generation quality.
\end{abstract}

\definecolor{initblue}{RGB}{189,215,238}   
\definecolor{flipone}{RGB}{91,155,213}     
\definecolor{fliptwo}{RGB}{31,78,121}    

\begin{figure}[t]
\begin{center}
\includegraphics[width=\columnwidth]{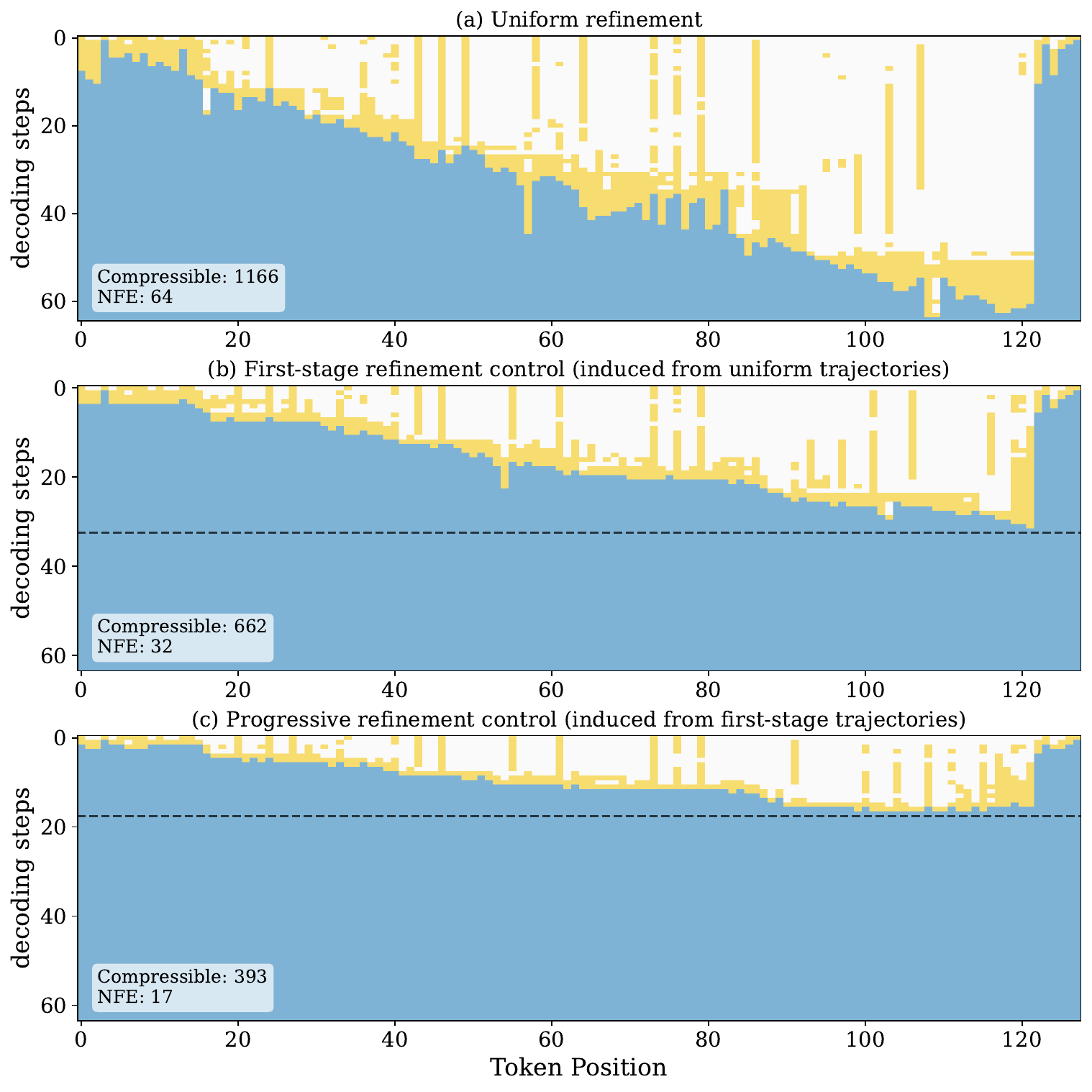}
\caption{Empirical visualization of refinement trajectory changes under refinement control.
Blue denotes unmasked tokens, while yellow marks redundant refinement where the current prediction already matches the final unmasked value; the dashed line indicates the final decoding step.
Starting from uniform top-2 refinement, inducing a first-stage refinement rule produces different refinement trajectories rather than merely shortening the refinement steps.
Further refinement control induced from the first-stage reshaped trajectories reorganizes the decoding trajectories again, demonstrating the dynamic nature of refinement control.
\label{fig:motivation}
}
\end{center}
\end{figure}

\section{Introduction}
Diffusion language models generate text through an iterative denoising process, gradually transforming an initial noisy sequence into a coherent output over multiple refinement steps \citep{sohl2015deep, ho2020ddpm, li2024scaling}.
Unlike autoregressive decoding, which factorizes generation into a left-to-right sequence of conditional distributions and decodes each token once it is produced, diffusion decoding predicts distributions over all positions at every refinement step, allowing token states to be repeatedly revised before unmasked \citep{austin2021structured, ye2025dream}. Consequently, decoders progressively unmask tokens whose predictions have converged.
This refinement paradigm enables flexible decoding orders and large-scale parallelism, naturally forming a dynamic refinement process over token-wise predictive states.

A central inefficiency of diffusion decoding lies in the use of a uniform refinement rule across all tokens \citep{ebsampler2025, prophet2025}.
In practice, different tokens stabilize at different rates, yet standard diffusion decoders apply the same refinement operator to all positions at every step.
As a result, substantial computation is spent repeatedly refining tokens that have already converged.
Whether a token has converged cannot be judged reliably from instantaneous uncertainty; it is determined by how its prediction changes over future refinement steps \citep{knowanswer2025, timeisafeature2025}.
In other words, the refinement trajectory provides the most direct supervision signal for refinement control.
Notably, refinement control does not operate on a fixed process: the refinement rule reshapes future refinement trajectories, which in turn determine how refinement rules should be formulated. As a result, refinement trajectories and refinement rules evolve together, making refinement control an inherently dynamic problem \citep{tauleaping1976}.
Fig.~\ref{fig:motivation} visualizes this redundancy and how refinement trajectories are progressively reshaped under refinement control.

Recent efforts approach refinement control by introducing auxiliary uncertainty-based signals to guide early stopping or token-level update decisions \citep{prophet2025, knowanswer2025, timeisafeature2025}. However, existing approaches assess refinement necessity from instantaneous or aggregated observations under a fixed process, neither treating refinement trajectories as supervision signals nor accounting for how refinement control reshapes the trajectories it induces.
In contrast, a temporal and dynamic refinement perspective is required that grounds refinement control in refinement trajectories and accounts for the fact that these trajectories are continuously revised under such control \citep{speculative2023, specdiff2025, spiffy2025}.

In this work, we introduce a progressive, trajectory-grounded refinement control framework. We derive a token-level notion of empirical convergence progress from full decoding trajectories, providing a continuous characterization of refinement necessity beyond instantaneous uncertainty. Based on this signal, we propose \emph{Progressive Refinement Regulation} (PRR), a lightweight token-wise controller that predicts convergence progress from decoding states. PRR regulates refinement through temperature-based distribution shaping, enabling converged tokens to be unmasked earlier while preserving refinement on unconverged ones \citep{li2024scaling, ye2025dream}.
Crucially, refinement control induces a supervision shift by changing the trajectories that define control supervision, which PRR addresses through a progressive self-evolving training scheme: at each stage, rollouts generated under the current PRR-regulated decoding process are used as a proxy for the refinement trajectories in the next training stage, and each update is regularized with a trust-region constraint that limits the change in token distributions between successive controllers. Experiments show that PRR substantially accelerates diffusion decoding across multiple benchmarks while preserving generation quality.

This paper makes three contributions.
First, we formulate diffusion decoding as a progressive refinement control problem over an induced, evolving refinement process, and introduce supervision shift as a central challenge \citep{surveyllm2023, zou2023survey_diffusion_nlp}.
Second, we introduce a notion of \emph{empirical convergence progress}, a temporal, token-level supervision signal derived from full decoding rollouts that characterizes refinement necessity from a trajectory perspective.
Third, we propose \emph{Progressive Refinement Regulation} (PRR), a refinement control approach that integrates trajectory-grounded supervision, progressive self-evolving training, and trust-region–constrained refinement regulation, substantially accelerating diffusion decoding while preserving generation quality.

\section{Background}
\subsection{Masked Discrete Diffusion Language Models}
Discrete diffusion language models formulate text generation as an iterative denoising process over sequences of categorical variables \citep{nie2025llada,ye2024beyondar,liu2024ddpd}. Given a fixed-length token sequence $\mathbf{x} = (x_1, \dots, x_L), ; x_i \in \mathcal{V}$, generation gradually transforms a noisy sequence into a coherent output over multiple refinement steps, rather than following a fixed autoregressive order.

A diffusion model defines a forward noising process and a corresponding reverse denoising process. In the discrete setting, a common instantiation is \emph{masked diffusion}, where noise is represented by a special absorbing symbol \texttt{[MASK]} \citep{chang2022maskgit,zheng2024timeagnostic,ou2024absorbing}. At refinement step $t$, the noised sequence $\mathbf{x}^{(t)}$ is obtained by independently replacing tokens in $\mathbf{x}^{(t-1)}$ with \texttt{[MASK]} according to a step-dependent masking probability:
\[
q\!\left(x_i^{(t)} \mid x_i^{(t-1)}\right) =
\begin{cases}
\texttt{[MASK]}, & \text{with probability } \alpha_t, \\
x_i^{(t-1)}, & \text{otherwise}.
\end{cases}
\]

The reverse process is parameterized by a neural network $p_\theta$ that predicts categorical distributions over the vocabulary for each position, conditioned on the partially masked sequence:
\[
p_\theta\!\left(\mathbf{x}^{(t-1)} \mid \mathbf{x}^{(t)}\right)
= \prod_{i=1}^L p_\theta\!\left(x_i^{(t-1)} \mid \mathbf{x}^{(t)}, t\right).
\]
During training, the model is optimized to recover tokens from a partially masked sequence at randomly sampled refinement steps. At inference time, decoding starts from a fully masked sequence and iteratively applies the reverse process for a fixed number of steps.

Unlike autoregressive models, diffusion models predict token distributions for all positions at every step, allowing token states to be revised multiple times before unmasked. Decoding therefore forms a dynamic refinement trajectory in which predictions may converge, or be revised over time. This iterative refinement enables flexible decoding orders, parallel updates, and revision of earlier predictions based on later context \citep{wang2023schedules,wang2025remdm}.

In the next section, we abstract this refinement trajectory into a discrete refinement process and formulate refinement control on top of it.

\subsection{Perspectives on Diffusion Decoding Control}
Recent diffusion language models have motivated growing interest in controlling the decoding process, with the goal of improving efficiency and accuracy at inference time \citep{nie2025llada,song2025seeddiffusion,khanna2025mercury}. 

A first class of approaches treats decoding control as a global scheduling problem, emphasizing \emph{when to stop} refinement. 
Under this view, decoding terminates once uncertainty- or confidence-based criteria indicate convergence, typically relying on step-wise or aggregated statistics over the entire sequence \citep{wei2025slowfast,israel2025apd,wang2023schedules}.
A second line of work formulates decoding control as a token selection problem, focusing on \emph{where to refine}. 
These methods update only subsets of positions at each step, often guided by confidence- or entropy-based signals, in order to organize parallel updates or reduce redundant computation \citep{huang2025ctrldiff,liu2024ddpd,kim2025tokenordering}.

A further line of work studies diffusion decoding control from a learning-based perspective, where learned policies or controllers adapt decoding behavior based on intermediate states \citep{huang2025ctrldiff,ye2024beyondar,wei2025slowfast}. 
In parallel, distillation-based approaches pursue acceleration by compressing the denoising process into faster models with fewer refinement steps \citep{zhu2025dimo,debortoli2025specdiffusion,gat2024dfm}.
A comprehensive discussion of related approaches is provided in Appendix~\ref{app:related} \citep{zhang2025surveyparallel}.

While these perspectives differ in their control abstractions, they are primarily formulated around global schedules, discrete token selection, or implicit uncertainty heuristics. 
They do not explicitly model whether individual tokens have entered and remain on their final refinement trajectories, nor how refinement control reshapes the refinement process that generates such trajectories. 
In contrast, we adopt a dynamical view of diffusion decoding and treat refinement as a non-stationary process whose dynamics evolve with the refinement rules. Under this view, whether further refinement is necessary for a token is defined by its future decoding trajectory under the induced process, rather than by instantaneous uncertainty or fixed global criteria.
This framing motivates a progressive, trajectory-conditioned refinement control mechanism, which we develop in the next section.

\section{Method}
\subsection{Refinement Control under Evolving Refinement Processes}
\label{sec:method:formulation}

\paragraph{Refinement dynamics.}
Following Section~2, discrete diffusion decoding produces predictive distributions
$p_\theta(x_i^{(t-1)} \mid \mathbf{x}^{(t)}, t)$ at each step \citep{sahoo2024masked, zheng2023reparam, hoogeboom2021ardm}.
We consider decoding as a finite-horizon refinement process over token-wise predictive states:
\begin{align*}
p_{i,t}(\cdot)& = p_\theta(x_i^{(t-1)} \mid \mathbf{x}^{(t)}, t) \in \Delta(\mathcal{V}),  \\ 
s_t& = (p_{1,t}, \dots, p_{L,t}), \qquad t \in \{1,\dots,T\}.
\end{align*}
The diffusion decoder induces a stochastic refinement process governed by a base refinement operator $\mathcal{F}_0$,
\[
s_{t+1} \sim \mathcal{F}_0(s_t, t).
\]
The resulting sequence $\{s_t\}_{t=1}^T$ 
forms a discrete-time, coupled refinement process.

\begin{figure*}[t]
\begin{center}
\includegraphics[scale=0.47]{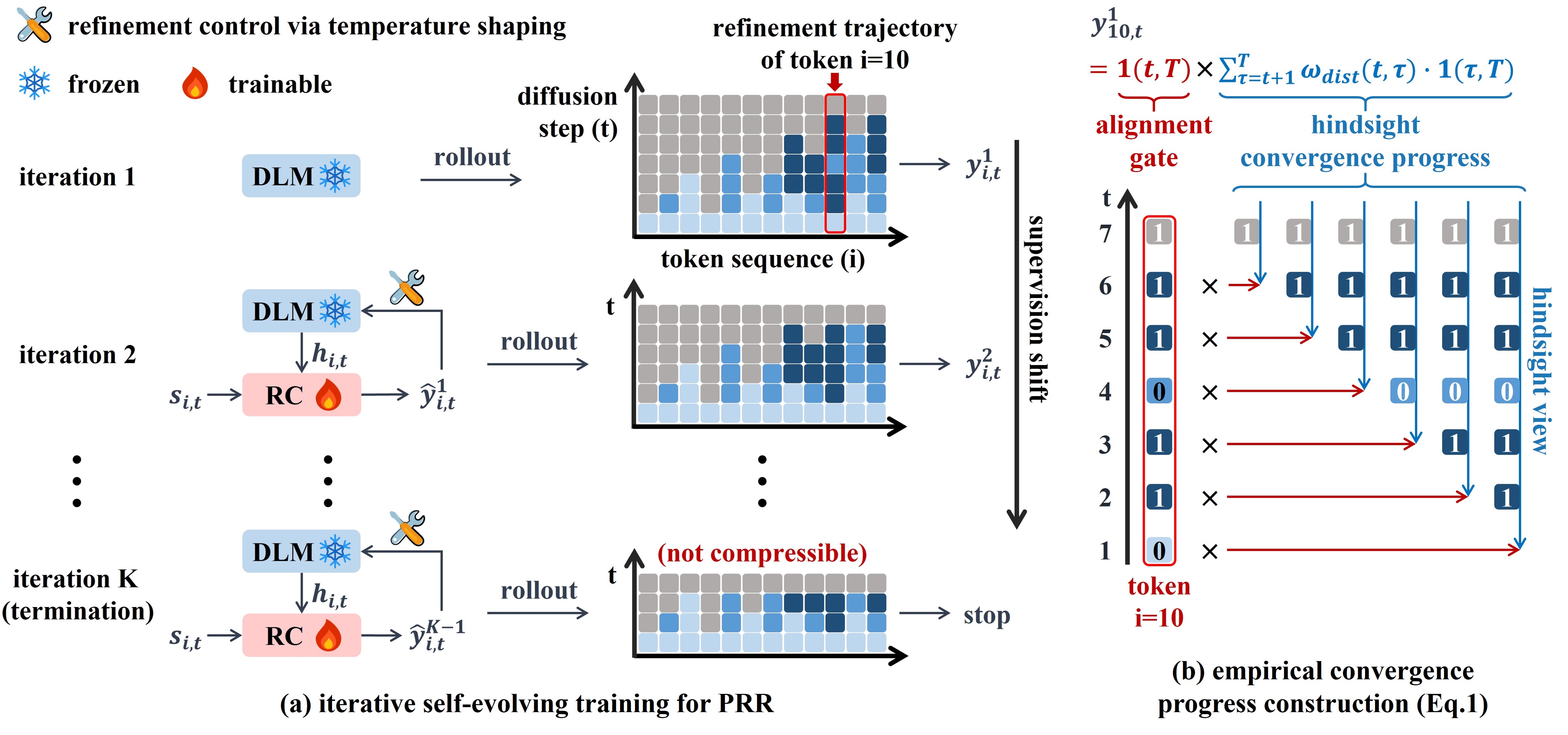}
\caption{
(a) \emph{Progressive self-evolving training of PRR.} 
At each stage, a refinement controller regulates diffusion decoding to induce refinement trajectories, which are then used to construct supervision for the next-stage controller.
(b) \emph{Empirical stability signal construction.}
For a given token, the refinement trajectory records its predictions across denoising steps.
The empirical stability target $y_{i,t}$ is computed from distance-weighted suffix agreement with the final decoded token (Eq.~1), quantifying whether the current prediction has aligned with the final outcome and how persistently this alignment holds.
}\label{fig:method}
\end{center}
\end{figure*}

\paragraph{Refinement Control.}
The base refinement operator applies a fixed refinement rule throughout decoding, even though different tokens exhibit different refinement progress over time, leading to redundant refinement steps.
Refinement control acts directly on the refinement dynamics.
We introduce a family of regulation operators $\mathcal{I}_\phi$ that transform the base dynamics, yielding
\[
s_{t+1} \sim \mathcal{F}_\phi(s_t, t) = \mathcal{I}_\phi(\mathcal{F}_0, s_t, t).
\]
Decoding under $\mathcal{F}_\phi$ induces a refinement trajectory
\[
\mathcal{T}_\phi = (s^{\phi}_1, \dots, s^{\phi}_T),
\]
with induced trajectory distribution $\mathbb{P}_\phi(\mathcal{T})$.

The future trajectory provides a natural source of hindsight supervision for refinement control \citep{liu2024planneddenoising}.
We formalize the necessity of further refinement through a trajectory-grounded quantity
\[
y_{i,t} = \Psi(\mathcal{T}_0, i, t),
\]
where $\mathcal{T}_0$ is a rollout under $\mathcal{F}_0$ and $\Psi$ summarizes how the prediction at position $i$ evolves over subsequent steps $(s_{t+1}, \dots, s_T)$.

\paragraph{Supervision shift under refinement control.}
Regulating the refinement dynamics changes the trajectory distribution generated by the decoder:
\[
\mathcal{F}_0 \;\xrightarrow{\;\mathcal{I}_\phi\;}\; \mathcal{F}_\phi
\quad\Longrightarrow\quad
\mathbb{P}_0(\mathcal{T}) \neq \mathbb{P}_\phi(\mathcal{T}),
\]
where $\mathbb{P}_\phi(\mathcal{T})$ is induced by $s_{t+1} \sim \mathcal{F}_\phi(s_t,t)$.
As a result, supervision signals derived from trajectories under $\mathcal{F}0$ no longer reflect the behavior induced by $\mathcal{F}\phi$, and refinement control cannot be treated as learning under a fixed trajectory distribution.
Instead, the supervision data evolves together with the refinement dynamics, making refinement control an inherently progressive problem.

\paragraph{Problem statement.}
We study refinement control in diffusion decoding as the problem of designing a regulation operator $\mathcal{I}_\phi$ that shapes the base refinement dynamics $\mathcal{F}_0$.
Supervision for refinement control is constructed from trajectories generated by the refinement dynamics.
As refinement control changes the dynamics, it also changes the trajectories that provide supervision, inducing a progressive refinement process.

\subsection{Trajectory-Grounded Convergence Progress}
\label{sec:method:trajectory_signal}
Based on the formulation in Section~3.1, we now specify the construction of the trajectory-grounded convergence progress
$y_{i,t} = \Psi(\mathcal{T}_0, i, t)$,
which characterizes whether token $i$ still requires refinement at step $t$.
Our goal is to derive a continuous, token-level supervision signal from refinement trajectories 
that reflects whether a token’s current prediction has aligned with the final decoded outcome and how persistently this alignment holds over subsequent refinement steps.
Such information is not reliably accessible from instantaneous uncertainty measures alone, since a token’s apparent confidence at a given step does not reveal whether it will later be revised.

Given an input prompt, we run the base diffusion decoder for $T$ steps and record the token-level predictive distributions
$\{p_{i,t}\}_{t=1}^{T}$.
Let $\hat{y}_{i,t} = \arg\max_v p_{i,t}(v)$ denote the top-probability token at position $i$ and step $t$, and let $\hat{y}_{i,T}$ denote the final decoded token.
A rollout under the base refinement operator $\mathcal{F}_0$ produces a refinement trajectory $\mathcal{T}_0 = (s_1,\dots,s_T)$, which defines a discrete refinement path for each token across refinement steps.

\paragraph{Empirical convergence progress.}
For each token $i$ and step $t$, we define a trajectory-grounded supervision signal
$y_{i,t} \in [0,1]$ as
\begin{equation}
\label{eq:yemp}
y_{i,t}
=
\mathbf{1}\!\left(\hat{y}_{i,t} = \hat{y}_{i,T}\right)
\cdot
\sum_{\tau = t+1}^{T}
\omega_{\text{dist}}(t,\tau)
\cdot
\mathbf{1}\!\left(\hat{y}_{i,\tau} = \hat{y}_{i,T}\right),
\end{equation}
where $\mathbf{1}(\cdot)$ denotes the indicator function.
The leading indicator acts as a gate: if the current prediction has not yet aligned with the final decoded token, the entire signal is 0.
When this gate is active, the weighted suffix term measures how consistently the token remains aligned with the final outcome over subsequent refinement steps.
This construction maps a full refinement trajectory to a dense, continuous token-level signal that reflects whether a token has entered its final refinement path and how persistently it remains on that path.

The weights $\omega_{\text{dist}}(t,\tau)$ determines the contribution of each future step.
In this work, we adopt a simple linear decay that emphasizes closer future refinement steps.
\begin{equation}
\label{eq:wdist}
\omega_{\text{dist}}(t,\tau)
=
\frac{T - \tau + 1}{\sum_{u = t+1}^{T} (T - u + 1)}.
\end{equation}
This assigns larger weight to closer future steps and gradually decreases the contribution of more distant ones, with normalization
$\sum_{\tau = t+1}^{T} \omega_{\text{dist}}(t,\tau) = 1$.

Overall, $y_{i,t}$ can be interpreted as a distance-weighted suffix consistency score:
it becomes nonzero once a token aligns with the final decoded outcome, and increases as the token remains more persistently aligned with the final outcome over subsequent refinement steps.

\begin{figure*}[ht]
\begin{center}
\includegraphics[width=1.91\columnwidth]{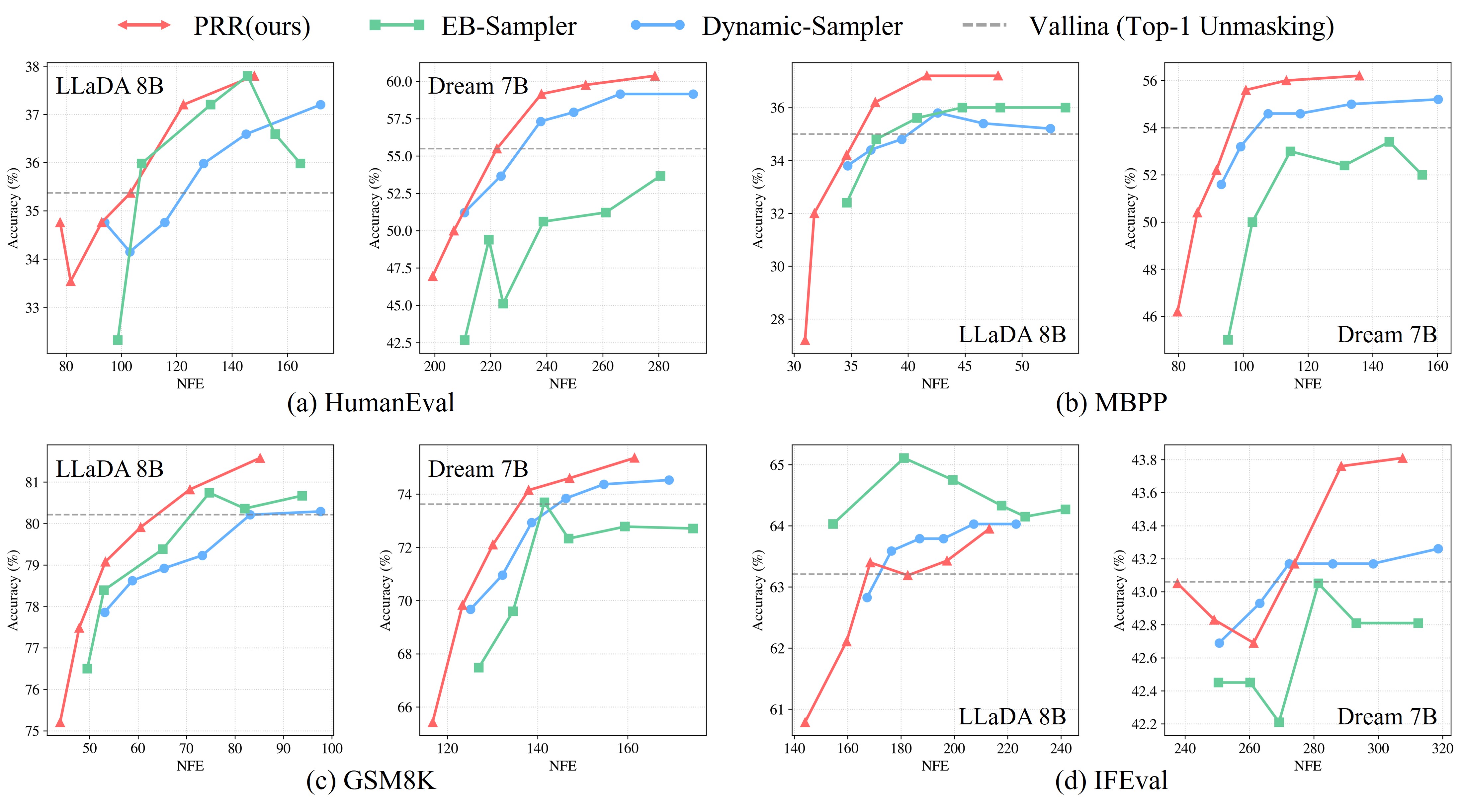}
\caption{Accuracy versus number of function evaluations (NFE) on HumanEval, MBPP, GSM8K, and IFEval under Dream-7B and LLaDA-8B backbones.}
\label{fig:tradeoff}
\end{center}
\end{figure*}

\subsection{Progressive Refinement Control via Trust-Region Regulation}
\label{sec:method:control}
We now introduce \emph{Progressive Refinement Regulation} (PRR), a refinement control framework for diffusion decoding under supervision shift (Section~3.1).
At progressive stage $k$, we generate rollouts $\mathcal{T}_{\phi_k}$ using the controller $\phi_k$ and construct supervision signals $y^k_{i,t}$ following Section~3.2, which are then used to train the next-stage controller $\phi_{k+1}$.
Figure~\ref{fig:method} provides an overview of this progressive self-evolving training paradigm and the construction of the empirical convergence progress signal.

\paragraph{Progressive refinement regulator.}
We instantiate refinement regulation through a lightweight regulator $g_\phi$ that predicts a token-wise refinement necessity from the instantaneous state:
\begin{equation}
\label{eq:controller}
\hat{y}_{i,t} = g_\phi(s_t, t, i) \in \mathbb{R},
\end{equation}
which estimates the empirical refinement progress $y_{i,t}$ defined in Section~3.2.
Here $s_t$ denotes the refinement state, constructed from the denoising model’s hidden representations together with a small set of refinement features such as token entropy, global unmask rate, and the diffusion step.

The controller output is used to regulate the sharpness of the predictive distribution.
Specifically, we transform the base predictive distribution $p_{i,t}(\cdot)$ using temperature regulation \citep{hinton2015distill, guo2017calibration}:
\begin{equation}
\label{eq:temp_mod}
p'_{i,t}(v)
=
\frac{p_{i,t}(v)^{1 / \tau_{i,t}}}{\sum_{u \in \mathcal{V}} p_{i,t}(u)^{1 / \tau_{i,t}}}.
\end{equation}
where $\tau_{i,t} = h(\hat{y}_{i,t})$ maps the controller output to a positive temperature.
In our implementation, we use a simple linear mapping
\begin{equation}
\tau_{i,t} = \tau_0 \big(1 + \alpha \cdot \hat{y}_{i,t}\big),
\label{eq:tau_alpha}
\end{equation}
where $\hat{y}_{i,t} \in [0,1]$ denotes the predicted empirical refinement progress for token $i$ at step $t$, $\tau_0$ is a fixed base temperature, and $\alpha$ controls the strength of refinement regulation.
When $\alpha=0$ (and $\tau_0=1$), Eq.~\eqref{eq:tau_alpha} reduces to the vanilla diffusion decoding rule.
Larger $\tau_{i,t}$ encourages flatter distributions and continued exploration, while smaller $\tau_{i,t}$ sharpens the distribution and accelerates stabilization.

Applying the decoding rule to $\{p'_{i,t}\}_{i=1}^L$ yields the next predictive state $s_{t+1}$ and defines the regulated refinement dynamics
\begin{equation}
s_{t+1} \sim \mathcal{F}_\phi(s_t,t).
\end{equation}

\paragraph{Trust-region constrained progressive regulation.}
Compared to one-shot training under a fixed trajectory distribution, this progressive training procedure alleviates the trajectory mismatch to some extent, 
because supervision is repeatedly constructed under the most recent refinement dynamics rather than from a single, fixed base process.
However, it does not eliminate supervision shift.
We therefore introduce a trust-region regularization to ensure that successive controllers induce smoothly varying token distributions \citep{schulman2015trpo, schulman2017ppo}:
\begin{align*} 
\label{eq:kl} 
\mathcal{L}(\phi_{k+1})& \;=\; 
\mathbb{E}_{(s_t,t)\sim \mathcal{T}_{\phi_k}} 
\Big[\ell\!\big(g_{\phi_{k+1}}(s_t,t,i),\; y^{k}_{i,t}\big)\Big] 
\\ 
+&\lambda \, 
\mathbb{E}_{(s_t,t)\sim \mathcal{T}_{\phi_k},\, i \sim [L]} 
\Big[ \mathrm{KL}\!\big(p'_{i,t,\phi_k}(\cdot)\,\|\,p'_{i,t,\phi_{k+1}}(\cdot)\big)\Big], 
\end{align*}
where $y^{k}_{i,t}$ denotes the trajectory-grounded supervision constructed from rollouts under $\phi_k$, 
$p'_{i,t,\phi}(\cdot)$ denotes the temperature-shaped predictive distribution at token $i$ and step $t$ induced by controller $\phi$ (Eq.~\eqref{eq:temp_mod}), 
and the KL divergence is applied token-wise and step-wise.
This trust-region regularization constrains the drift of the induced refinement process and stabilizes progressive refinement regulation.

\paragraph{Inference-time refinement control.}
At inference time, we run diffusion decoding with the learned controller fixed.
At each step $t$, the controller predicts $\hat{y}_{i,t}=g_\phi(s_t,t,i)$ from the instantaneous predictive state, maps it to temperatures $\{\tau_{i,t}\}$, reshapes the token distributions via Eq.~\eqref{eq:temp_mod}, and applies the decoding rule to accelerate the refinement process.
This regulation suppresses redundant refinement updates on converged tokens while maintaining refinement on uncertain ones, enabling fewer effective refinement steps without sacrificing final quality.

\section{Experiments}
\subsection{Experimental Setup}
We conduct experiments on two recent discrete diffusion language models, LLaDA-1.5-8B-Base \citep{nie2025llada} and Dream-v0-Base-7B \citep{ye2025dream}, to evaluate Progressive Refinement Regulation (PRR).
All backbone models are kept frozen, and PRR trains a lightweight controller on top of them to regulate refinement strength during decoding.
We evaluate on a diverse set of reasoning and code-generation benchmarks, including GSM8K (5-shot) \citep{cobbe2021gsm8k}, HumanEval (0-shot) \citep{chen2021humaneval}, MBPP (3-shot) \citep{austin2021mbpp}, IFEval (0-shot) \citep{zhou2023ifeval}, and MATH (4-shot) \citep{hendrycks2021math}. 
We use generation length $L=256$ for GSM8K and $L=512$ for all other benchmarks.

Both LLaDA and Dream use block-wise decoding with block size $B=32$ \citep{blockdiffusion2023}.
Generation proceeds until all tokens in the sequence are fully unmasked, rather than terminating upon encountering an end-of-sequence (EOS) token. 
We evaluate both generation quality and decoding efficiency. 
Quality is measured by exact match accuracy under the standard answer extraction protocol for GSM8K and MATH, 
Pass@1 for HumanEval and MBPP, 
and instruction-level loose accuracy for IFEval. 
Decoding efficiency is quantified by the Number of Function Evaluations (NFE), defined as the total number of diffusion model forward passes invoked until the generation process terminates.

To isolate the effect of progressive refinement regulation from other decoding or acceleration mechanisms, 
we evaluate four decoding strategies, all built upon the dynamic-sampler framework \citep{wu2025fastdllm} and incorporating DualCache during inference:
\emph{Vanilla} (frozen backbone with the standard top-1 unmasking rule),
\emph{Dynamic-Sampler} (dynamic unmasking controlled by a confidence threshold),
\emph{EB-Sampler} (an entropy-based adaptive decoding baseline)~\citep{benhamu2025ebsampler},
and \emph{PRR} (our proposed progressive refinement regulation method).

Appendix~\ref{demo} visualizes and compares the refinement dynamics induced by PRR and the LLaDA (Vanilla) on GSM8K.

\begin{table*}[htbp]
\centering
\small
\setlength{\tabcolsep}{4pt}
\begin{tabular}{l|ccccc|ccccc}
\hline
\multirow{2}{*}{\textbf{Method}}
& \multicolumn{5}{c|}{\textbf{Accuracy} ($\uparrow$)}
& \multicolumn{5}{c}{\textbf{NFE} ($\downarrow$)} \\
& GSM8K & HumanEval & MBPP & IFEval & MATH
& GSM8K & HumanEval & MBPP & IFEval & MATH \\
\hline
\multicolumn{11}{c}{\textbf{Dream-7B}} \\
\hline
Vanilla          
& 73.62 & 55.49 & 54.00 & 43.06  & 38.82
& 256 & 512 & 512 & 512  & 512\\
Dynamic-Sampler        
& 72.93 & 57.32 & 54.60 & 43.26  & 38.62 
& 138.68 & \textbf{237.82} & 107.67 & 318.57 & 161.23\\
EB-Sampler        
& 73.69 & 50.61 & 53.00 & 43.05  & 38.30 
& 141.50 & 238.87 & 114.57 & \textbf{281.41} & 153.94\\
\textbf{PRR(Ours)}  
& \textbf{74.15} & \textbf{59.15} & \textbf{55.60} & \textbf{43.76}  & \textbf{39.02}
& \textbf{138.02} & 238.08 & \textbf{100.84} & 288.50 & \textbf{149.78}\\
\hline
\multicolumn{11}{c}{\textbf{LLaDA-8B}} \\
\hline
Vanilla             
& 80.21 
& 35.37
& 35.00   
& 63.21
& 33.58 & 256 & 512 & 512 & 512  & 512\\
Dynamic-Sampler       
& 79.23 & 35.98 & 35.80 & 63.59 & 33.30  
& 73.23 & 129.76 & 42.61 & 176.40 & 142.10\\
EB-Sampler  
& 80.74 & 37.20 & 36.00 & \textbf{65.11}  & 33.52
& 74.72 & 132.30 & 44.76 & 181.16  & 148.32\\
\textbf{PRR(Ours)}  
& \textbf{80.82} & \textbf{37.20}  & \textbf{37.20} & 63.40   & \textbf{33.94} 
& \textbf{70.69} & \textbf{122.43} & \textbf{41.65}  & \textbf{168.48} & \textbf{133.98}\\
\hline
\end{tabular}
\caption{Main results on Dream-7B and LLaDA-8B.Accuracy ($\uparrow$): higher is better. NFE ($\downarrow$): lower is better.
PRR consistently achieves higher accuracy under similar or lower decoding budgets, improving the accuracy-efficiency trade-off across reasoning and code-generation benchmarks.}
\label{tab:main_results}
\end{table*}

\subsection{Main Results: Accuracy-Efficiency Tradeoff}
Our main objective is to examine whether progressive refinement control can reshape the accuracy-efficiency tradeoff of diffusion decoding. 
To trace out this tradeoff, we sweep a confidence threshold for PRR: at each step, all tokens whose regulated confidence exceeds the threshold are unmasked. 
Varying this threshold from $0.7$ to $0.95$ yields a spectrum of decoding behaviors, where higher thresholds unmask fewer tokens per step and thus incur larger NFE. 
For the dynamic sampler, we similarly sweep a confidence threshold, but apply it to the raw model confidence before regulation. 
For EB-Sampler, we sweep the entropy bound parameter $\gamma$, which controls how many tokens can be unmasked per step.

\paragraph{Accuracy-efficiency frontiers.}
Figure~\ref{fig:tradeoff} shows that PRR shifts the accuracy--efficiency frontier upward and across most benchmarks on both backbones.
However, the improvement is not uniform: on LLaDA-8B evaluated on IFEval, PRR is inferior to EB-Sampler in the higher-NFE regime.
These results indicate that the benefit of refinement regulation is task- and backbone-dependent.

\paragraph{Accuracy at approximately matched NFE.}
Table~\ref{tab:main_results} reports accuracy at comparable decoding budgets.
On \textbf{Dream-7B}, PRR is the top-performing method on all five benchmarks, outperforming vanilla decoding, the dynamic sampler, and EB-Sampler under similar NFEs.
On \textbf{LLaDA-8B}, PRR improves over vanilla decoding on every benchmark and exceeds the dynamic sampler on four out of five tasks, with the only exception being IFEval where the dynamic sampler is slightly higher.
Compared to EB-Sampler, PRR achieves higher accuracy on three benchmarks, while EB-Sampler is better on HumanEval and IFEval.
Overall, the table highlights that PRR's gains are broad but not uniform: it consistently improves matched-budget accuracy on most tasks.

\begin{figure}[ht]
\begin{center}
\includegraphics[width=\columnwidth]{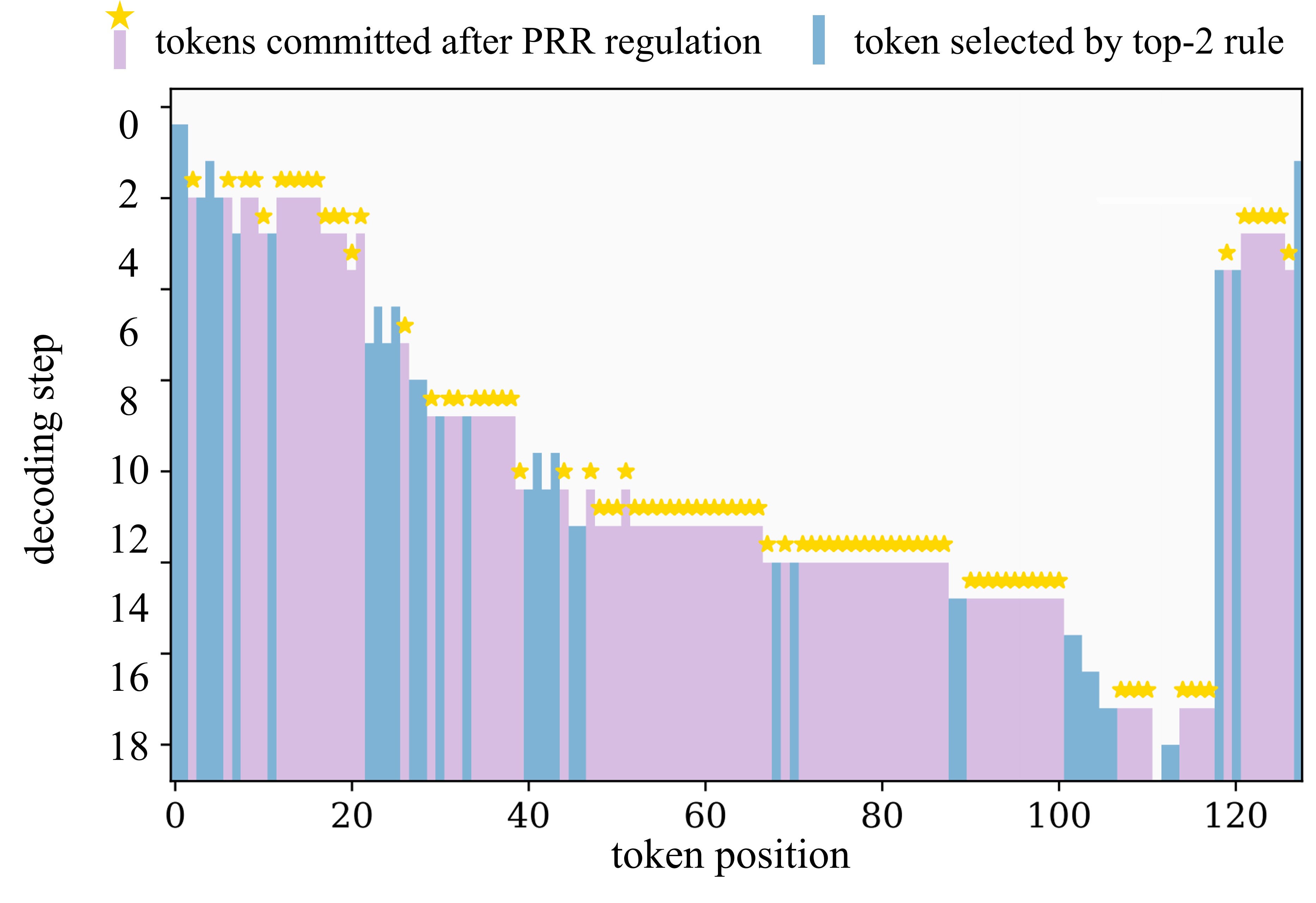}
\caption{
Token-level unmasking schedule during PRR-regulated decoding on one example.
At each decoding step, blue bars mark tokens selected by the top-1 refinement rule, while purple bars indicate additional tokens refined under PRR’s regulation (i.e., the expanded unmasking set after regulation). Yellow stars denote the positions newly unmasked at each step.
}
\label{fig:accelerate}
\end{center}
\end{figure}

\begin{figure*}[ht]
\begin{center}
\includegraphics[width=1.9\columnwidth]{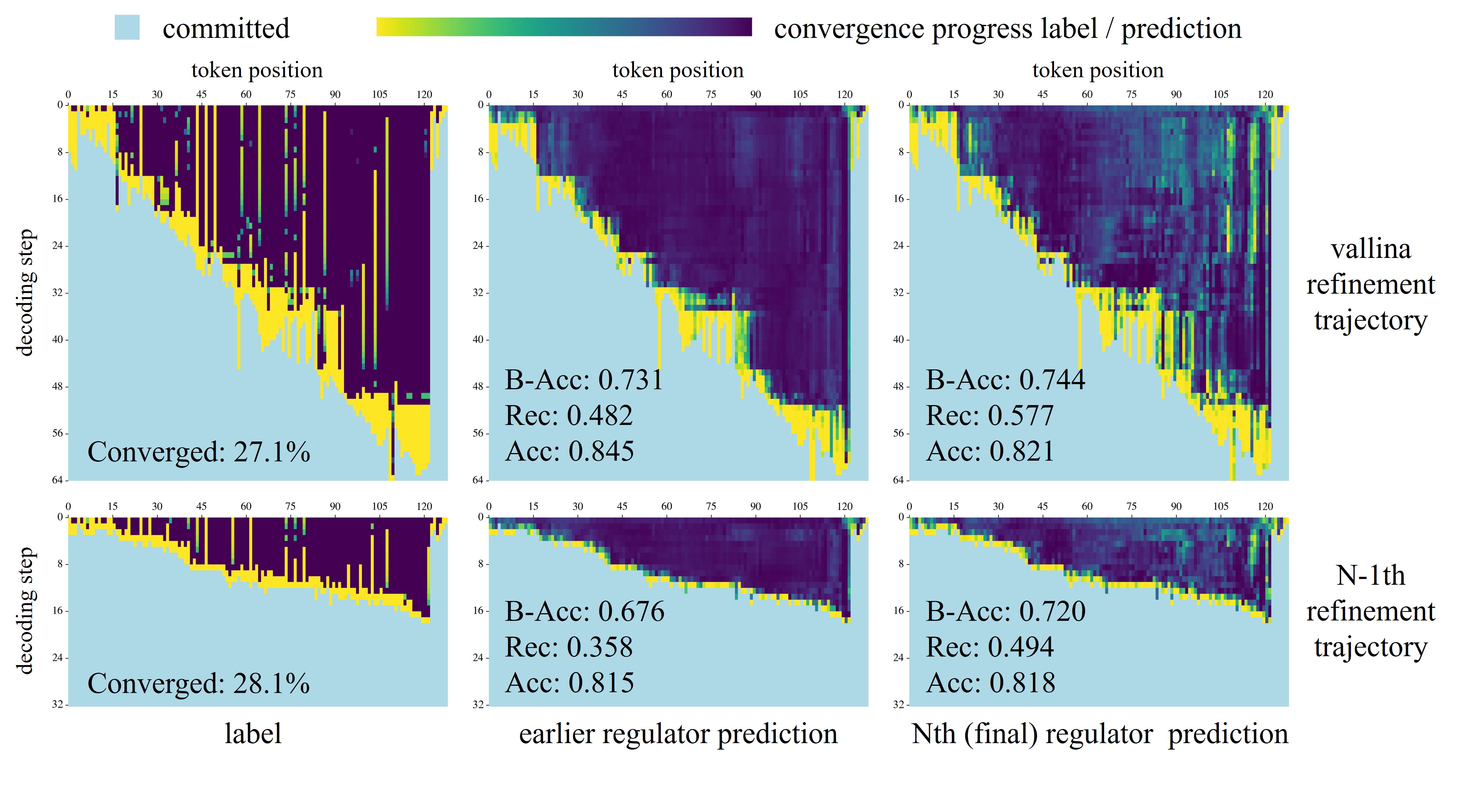}
\caption{Trajectory-grounded convergence progress labels and controller predictions.
Left: empirical convergence progress labels derived from full decoding rollouts (Eq.~(1)).
Light-blue masks denote positions that have already been decoded, and the colormap shows convergence progress over positions that remain under refinement.
Middle: predictions of the early-stage controller.
Right: predictions of the final controller.
The top row corresponds to vanilla refinement trajectories, while the bottom row shows PRR-regulated trajectories from the $(N\!-\!1)$-th training stage.
Panel annotations report decoded-token ratios and prediction metrics over the token-step grid.}
\label{fig:control_label}
\end{center}
\end{figure*}
\subsection{Token-Level Unmasking Scheduling under PRR}
\label{sec:token_schedule}

While the main results show that PRR substantially reduces inference steps, it remains unclear how this acceleration is realized along the decoding trajectory.
We therefore visualize the token-level unmasking process of PRR on a representative example to examine how per-step unmaskings are redistributed over positions.

Figure~\ref{fig:accelerate} visualizes how PRR regulates the per-step unmasking set during decoding.
PRR’s unmasking decisions are spatially clustered, rather than uniformly spread across positions.
Once a subset of tokens within a contiguous region has been unmasked, PRR tends to advance the entire region together, while continuing to refine a small set of hard positions.
This observation shows that PRR reshapes decoding at the granularity of individual token positions, inducing a structured unmasking process instead of uniformly advancing all positions.
Such token-level unmasking scheduling provides a concrete explanation for how PRR reduces redundant refinement steps without truncating the overall decoding procedure.

\subsection{Predicting Empirical Convergence Progress}
\label{sec:control_label}
PRR departs from existing decoding strategies by explicitly regulating whether a token still requires refinement.
This relies on two central questions: (1) whether empirical convergence progress is a trajectory-level property that can be inferred from intermediate diffusion states, and (2) whether such predictions remain accurate under the supervision shift induced when decoding trajectories are reshaped by PRR.
Figure~\ref{fig:control_label} visualizes the trajectory-grounded convergence progress labels and the corresponding PRR controller predictions on the same decoding example, evaluated on both vanilla and PRR-regulated trajectories.

\paragraph{Learning trajectory-grounded supervision.}
On vanilla decoding trajectories (top row), the early-stage controller already achieves non-trivial performance, with a balanced accuracy of $0.731$ and a recall of $0.482$ on unconverged regions.
After progressive training, balanced accuracy improves to $0.744$ and recall to $0.577$, indicating substantially better coverage of tokens that still require refinement, with only a minor drop in overall accuracy ($0.845 \rightarrow 0.821$).

On PRR-regulated trajectories (bottom row), the gains are more pronounced.
Balanced accuracy increases from $0.676$ to $0.720$, and recall from $0.358$ to $0.494$, while overall accuracy remains nearly unchanged ($0.815 \rightarrow 0.818$).
These results indicate that empirical convergence progress is predictable from intermediate diffusion states, and that progressive training significantly improves the controller’s ability to identify tokens that have not yet converged.

\paragraph{Progressive self-evolution under supervision shift.}
The early-stage controller performs worse on PRR-regulated than vanilla trajectories (e.g., $0.676$ vs.\ $0.731$ accuracy; $0.358$ vs.\ $0.482$ recall), reflecting a mismatch between its training distribution and the induced trajectories.
After progressive self-evolution, improvements are substantially larger on PRR-regulated than vanilla trajectories (e.g., $+0.044$ vs.\ $+0.013$ accuracy; $+0.136$ vs.\ $+0.095$ recall).

This asymmetric improvement provides direct evidence of the supervision shift introduced by PRR, and shows that progressive self-evolution is necessary for the controller to remain accurate as decoding trajectories are reshaped.
Further ablations on refinement regulation and progressive self-evolution are reported in Appendix~\ref{app:ablation}, and additional analyses of training dynamics and long-horizon stability are provided in Appendix~\ref{app:dynamics}.

\section{Conclusion}
We studied diffusion language model decoding from the perspective of \emph{progressive refinement control}, formulating refinement as a dynamic control problem over evolving refinement trajectories rather than a fixed denoising procedure. From this view, whether a token should continue to be refined cannot be determined from instantaneous signals under a fixed process, but is more naturally defined in terms of its refinement trajectory. Moreover, refinement control reshapes decoding itself, inducing a supervision shift where the trajectories that define control supervision are continuously altered.

We introduced \emph{empirical convergence progress}, a trajectory-grounded token-level signal derived from full decoding rollouts, and proposed \emph{Progressive Refinement Regulation} (PRR), a lightweight controller trained through progressive self-evolution to regulate refinement under this supervision shift. Experiments show that PRR improves the accuracy-efficiency tradeoff of diffusion decoding across multiple benchmarks, and our analysis demonstrates that empirical convergence progress is predictable from intermediate states and that progressive self-evolution is necessary for reliable refinement control.

\section{Impact Statement}
This paper studies diffusion language model decoding from the perspective of progressive refinement control, reframing decoding as a dynamic process whose trajectories evolve under refinement rules. By grounding refinement decisions in token-level refinement trajectories, the work offers a new viewpoint on how convergence and redundancy can be characterized and exploited during diffusion decoding.

This perspective may inspire new approaches to diffusion model acceleration and decoding control that move beyond fixed schedules or step-wise heuristics, potentially enabling more efficient and adaptive generation.

\bibliography{example_paper}
\bibliographystyle{icml2026}

\appendix

\section{Relate Works}\label{app:related}
\subsection{Diffusion Decoding Schedules and Early-Exit Criteria}
Several works formulate diffusion decoding control as a stopping problem, where the denoising process terminates once predictions satisfy a stability criterion.
These approaches typically rely on training-free rules defined over confidence, margin, or uncertainty statistics and operate at a global or coarse-grained level.

SchED proposes a progress-aware early-exit strategy that compares aggregated confidence signals against a stage-dependent threshold \citep{mohamed2025sched}.
Prophet observes that correct answers can often be identified before the final denoising step and applies a margin-based criterion to decide when to stop refinement \citep{knowanswer2025}.
Early exiting has also been explored in other diffusion settings, where stopping is triggered once uncertainty metrics stabilize across steps \citep{moon2024earlyexit, wang2023schedules}.

Early-exit mechanisms further appear in structured decoding scenarios, such as in-place prompting, where confidence signals determine termination for different output components \citep{shen2025thinking}.
Refinement convergence has also been assessed using distributional distance criteria, which signal stopping when belief distributions become sufficiently close \citep{zhang2025latentrefinement}.

\textbf{Discussion.}
These approaches treat decoding control primarily as a global stopping decision governed by instantaneous or aggregated uncertainty signals.
While effective for determining when decoding can terminate, they abstract away the heterogeneous and coupled evolution of individual tokens.
In contrast, our work adopts a token-specific, trajectory-conditioned view, where refinement decisions depend on a token’s future evolution rather than a single global stopping criterion.

\subsection{Confidence- and Entropy-Based Decoding Planners}
Another line of work formulates diffusion decoding control as a \emph{where-to-update} problem, where a planner selects subsets of positions to unmask or refine at each denoising step based on instantaneous uncertainty signals such as confidence or entropy.
This perspective underlies common unmasking and remasking strategies in large discrete diffusion models, which iteratively update selected positions while leaving others masked \citep{gong2025diffucoder,nie2025llada,ye2025dream}.

Beyond greedy per-step selection, planner-style methods design structured update sets to increase parallelism without fully unmasking all positions.
DUS introduces a purely inference-time scheduler that partitions positions into dilated groups for parallel updates \citep{luxembourg2025dus}, while Lookahead Unmasking evaluates candidate unmasking paths using path-level uncertainty estimates \citep{park2025lookum}.

\textbf{Discussion.}
These approaches control decoding through discrete token selection, determining \emph{where} refinement is applied based on step-wise uncertainty.
However, they do not model how strongly a selected token should be refined, nor how its refinement state may evolve due to token interactions later in the trajectory.
In contrast, our work focuses on token-specific refinement dynamics, conditioning refinement strength on anticipated future evolution rather than step-wise selection decisions.

\subsection{Trajectory- and History-Aware Refinement Signals}
Several works leverage history- or trajectory-level signals across denoising steps to guide diffusion decoding, instead of relying solely on instantaneous uncertainty statistics.
The common motivation is that early-step predictions may fluctuate, and that cross-step information can provide more reliable indicators of refinement stability.

MDPO and its Running Confidence Remasking (RCR) variant use confidence statistics accumulated along the denoising trajectory to inform remasking decisions, emphasizing trajectory-aware criteria over single-step signals \citep{he2025mdpo}.
Other approaches analyze temporal consistency of intermediate predictions, using agreement or stability across steps as signals to assess semantic reliability during decoding \citep{wang2025timefeature, ou2024absorbing}.

Beyond test-time signals, trajectory-level information has also been incorporated into training objectives.
Some methods exploit cross-step consistency as auxiliary supervision or reinforcement targets, encouraging stable refinement trajectories without explicitly learning decoding policies \citep{yang2025consistencyrl}.

\textbf{Discussion.}
These approaches treat trajectory-level information primarily as signals for evaluating refinement stability, either to guide heuristic decisions or to regularize training.
Such signals are typically consumed implicitly and do not specify how refinement strength for an individual token should be regulated as decoding progresses.
In contrast, our work conditions token-specific control directly on anticipated future trajectory evolution, rather than relying on post hoc stability assessment.

\subsection{Learned Decoding Policies and Controllers}
Another line of work formulates diffusion decoding control as a learning problem, where control decisions such as masking, unmasking, or revising tokens are produced by a learned policy or controller conditioned on intermediate decoding states \citep{chen2024dpad}.

A representative instance is \emph{Learning Unmasking Policies for Diffusion Language Models}, which formalizes the decoding process as a Markov decision process and uses reinforcement learning to train token unmasking policies \citep{2025learningunmask}.
Similarly, SPG proposes a policy-gradient-based method tailored for masked diffusion models to optimize learned decoding strategies \citep{wang2025spg}.

Beyond pure RL optimization, learned policies have also been explored under regulated objectives.
For example, \emph{Improving Discrete Diffusion Unmasking Policies Beyond Explicit Reference Policies} studies KL-regularized MDP optimization for decoding policies with convergence guarantees \citep{hong2025klpolicy}, and \emph{dUltra} learns an unmasking planner jointly with the base diffusion model using on-policy reinforcement signals \citep{chen2025dultra}.
Guided Star-Shaped Masked Diffusion (G-Star) further introduces a learned scheduler that adapts token revision decisions based on intermediate decoding states, albeit outside an explicit RL framework \citep{meshchaninov2025gstar}.

\textbf{Discussion.}
These approaches define decoding control through learned policies over discrete actions, such as selecting which tokens to mask, unmask, or revise at each step.
While adaptive, this formulation primarily focuses on \emph{which} tokens to update, rather than how the refinement strength of an individual token should be regulated as decoding progresses.
In contrast, our work does not learn a policy over discrete decoding actions; instead, it learns token-specific refinement control conditioned on anticipated future evolution, decoupling refinement strength regulation from discrete token selection.

\subsection{Distillation for Accelerating Diffusion Decoding}
Another related direction accelerates diffusion decoding through distillation, with the goal of reducing the number of denoising steps required at inference time.
Rather than modifying decoding policies or control signals, these methods compress the denoising process itself by transferring knowledge from a multi-step teacher to a faster student.
Self-Distillation Through Time (SDTT) distills denoising trajectories of a diffusion language model into a student that operates with significantly fewer refinement steps \citep{deschenaux2024sdtt, zhu2025dimo}.

Beyond discrete language models, step-reduction and process-level distillation have also been explored in continuous diffusion settings, where multi-step generative processes are compressed into fewer iterations or single-step predictors \citep{salimans2022progressive, song2023consistency}.
These methods define acceleration at the level of the generative process or model parameters, rather than through adaptive control during decoding \citep{luo2023diffusiondistill}.

\textbf{Discussion.}
Distillation-based approaches achieve acceleration by modifying or replacing the underlying denoising process, effectively reducing the number of refinement steps executed at inference time.
In contrast, our work preserves the original diffusion process and instead focuses on learning token-level refinement control within decoding, reallocating refinement effort without compressing the generative trajectory itself.

\section{Implementation Details}
\label{app:implementation}
Here we provide additional implementation details on controller training, architecture, and optimization.

\paragraph{Diffusion decoding configuration.}
During training, we fix the generation length to $L=256$ and use 256 diffusion steps, with block size $B=32$. 
The same block-wise diffusion decoding framework is used throughout \citep{blockdiffusion2023}, and PRR is integrated without modifying the underlying denoising process.

\paragraph{Progressive training protocol.}
Rollout prompts are sampled from the training splits of a weighted mixture of instruction and reasoning datasets, including GSM8K, MATH, MBPP, and the \textsc{Tulu-3-SFT} mixture. 
We construct a prompt pool of approximately 20{,}000 training examples, and sample from it with weights roughly proportional to GSM8K:MATH:MBPP:Tulu-3-SFT = 1:1:0.25:1.
Evaluation benchmarks are strictly held out and are not used during controller training.
At each stage, 200 rollouts are collected under the current refinement controller to construct trajectory-grounded supervision. 
The controller is trained for 10 epochs using AdamW \citep{loshchilov2019adamw} with learning rate $1\times10^{-4}$ and batch size 320.

The unmasking confidence threshold is linearly annealed from 0.95 to 0.80 across stages. 
Unless otherwise specified, we use the temperature-based refinement regulation strategy with $\alpha=1.0$.

\paragraph{Refinement controller architecture.}
The refinement controller is a lightweight MLP operating on token-wise hidden states augmented with auxiliary refinement features. 
For LLaDA, the input dimension is $4096 + 11 = 4107$, and for Dream it is $3584 + 11 = 3595$. 
The 11 auxiliary features include global and block-level refinement statistics as well as local predictive features such as top-1 probability, margin, and entropy.

The controller consists of a LayerNorm \citep{ba2016layernorm}, a linear projection to 1024 hidden units, GELU activation \citep{hendrycks2016gelu} with dropout, one residual MLP block, and a final linear projection to a scalar output predicting the refinement control score.

\paragraph{Loss and trust-region regularization.}
The controller is trained using a masked binary cross-entropy loss against trajectory-grounded empirical refinement progress targets. 
To stabilize progressive training, we introduce a trust-region regularization term that penalizes the KL divergence between successive temperature-regulated token distributions induced by the controller.
We use a Huber loss form for this regularization with weight $\lambda = 3.0$.

\paragraph{Inference-time refinement control.}
At inference time, the controller regulates refinement by modulating the token-wise temperature of the diffusion model.
Given the controller output $h_{i,t}$ for token $i$ at step $t$, we map it to a temperature value
\[
\tau_{i,t} = \tau_0 (1 + \alpha \cdot h_{i,t}),
\]
where $h_{i,t} \in [0,1]$ denotes the normalized controller output and $\tau_0$ is the base temperature.
We then apply temperature-based reshaping to the predicted token distribution:
\[
p'_{i,t}(x) \propto p_{i,t}(x)^{1 / \tau_{i,t}}.
\]
This temperature-regulated distribution is used for confidence estimation and unmasking decisions.
Unless otherwise specified, we set $\alpha = 1.0$ and $\tau_0 = 1.0$.

\paragraph{Efficiency engineering.}
For Dream experiments, we employ a dual-cache mechanism during rollout collection and evaluation to reduce GPU memory usage and improve inference throughput, following prior diffusion LLM acceleration work \citep{dKVCache2025, dLLMCache2025}.

\section{Training Dynamics and Stability of Progressive Self-Evolution}
\label{app:dynamics}
\begin{figure}[h]
\begin{center}
\includegraphics[width=\columnwidth]{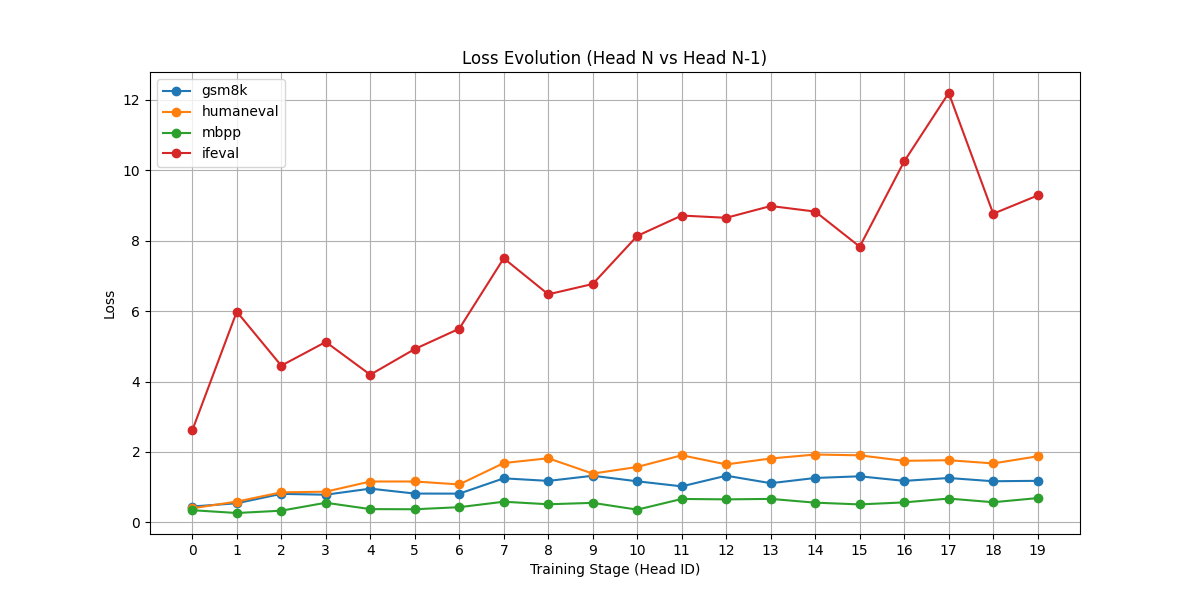}
\caption{
Training loss evolution across successive refinement heads.
Training loss of the controller across refinement heads on four benchmarks.
Loss increases as supervision is drawn from increasingly self-induced refinement trajectories, while remaining bounded across stages without divergence.
}
\label{fig:app_loss}
\end{center}
\end{figure}

\paragraph{Loss dynamics under self-induced distribution shift.}
Figure~\ref{fig:app_loss} shows the training loss across successive refinement heads.
Loss exhibits a clear upward trend, particularly on IFEval and HumanEval, reflecting the increasing difficulty of predicting empirical refinement progress on trajectories reshaped by earlier controllers.
Importantly, despite this growing distribution shift, loss remains bounded and does not diverge, indicating that progressive self-evolution increases supervision difficulty without destabilizing training.

\begin{figure}[h]
\begin{center}
\includegraphics[width=\columnwidth]{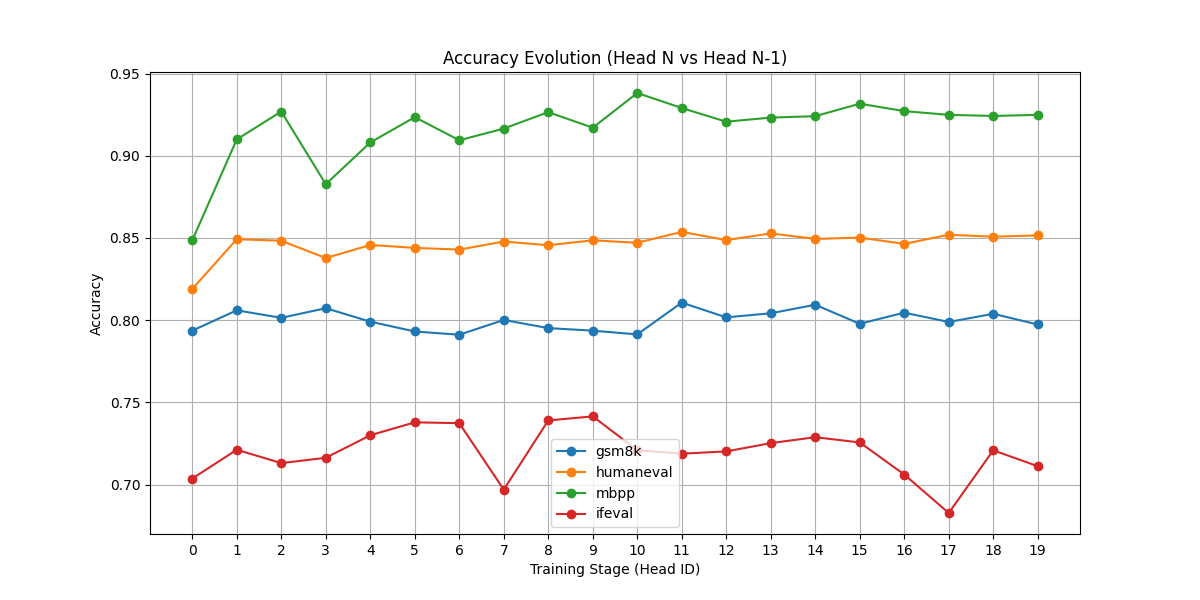}
\caption{
Accuracy evolution across successive refinement heads.
Controller accuracy remains stable across training stages, with only mild fluctuations.
}
\label{fig:app_accuracy_evolution}
\end{center}
\end{figure}

\paragraph{Absence of error accumulation.}
Figure~\ref{fig:app_accuracy_evolution} reports controller accuracy across refinement heads.
Across all benchmarks, accuracy remains stable with only minor non-monotonic fluctuations.
No systematic degradation is observed, indicating that regulators trained on self-generated trajectories do not accumulate prediction errors over stages.

\begin{figure}[h]
\begin{center}
\includegraphics[width=0.8\columnwidth]{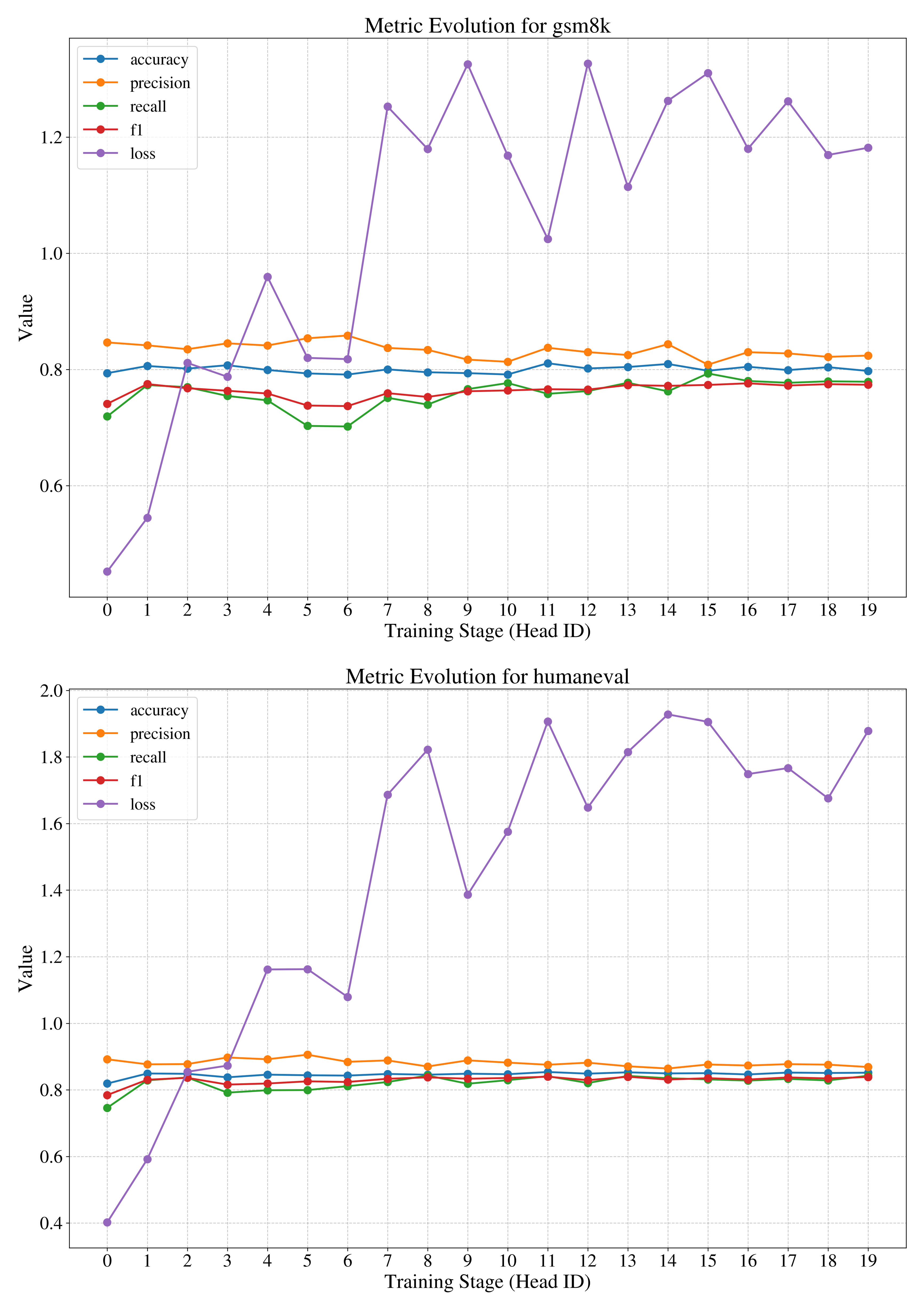}
\caption{
Training dynamics of the progressive self-evolving regulator on GSM8K and HumanEval.
We report accuracy, precision, recall, F1, and loss across successive training stages.
While loss increases substantially, all predictive metrics remain stable.
}
\label{fig:app_training_dynamics}
\end{center}
\end{figure}

\paragraph{Decoupling between loss growth and functional stability.}
Figure~\ref{fig:app_training_dynamics} shows detailed training dynamics on GSM8K and HumanEval.
As self-evolution proceeds, loss rises markedly, confirming that trajectory-grounded refinement supervision becomes progressively harder as refinement trajectories shift.
However, all task-level metrics (accuracy, precision, recall, F1) remain tightly concentrated.
This decoupling indicates that progressive self-evolution reshapes the refinement distribution without degrading regulator reliability, and that later-stage controllers adapt to their own induced trajectories rather than overfitting to early refinement patterns.

\section{Ablation Studies} 
\label{app:ablation}
\subsection{Progressive Self-Evolution and Trust-Region Regularization}
\label{sec:progressive_ablation}

\begin{figure}[t]
\begin{center}
\includegraphics[width=\columnwidth]{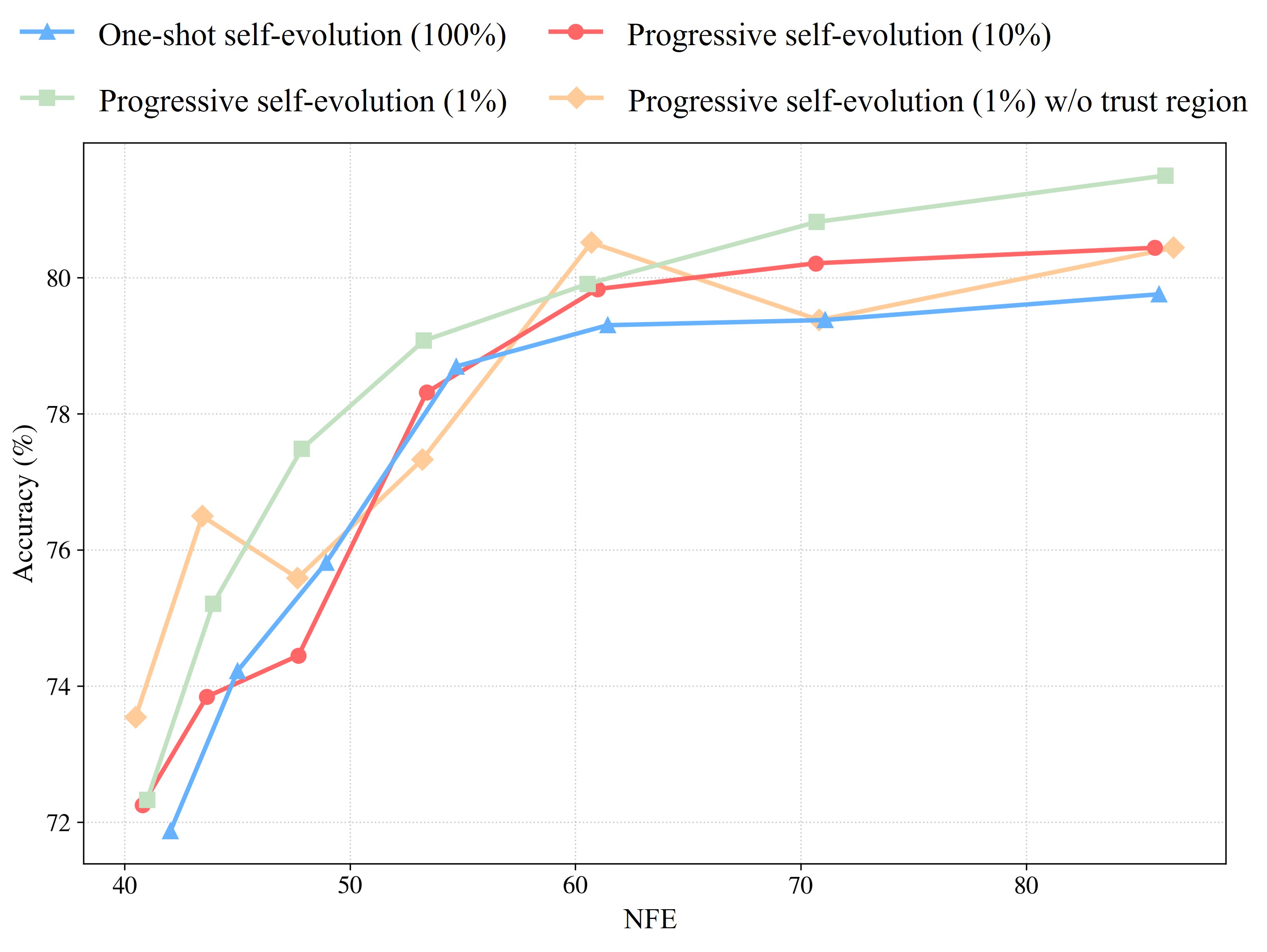}
\caption{
Effect of progressive self-evolution and trust-region regularization.
We report accuracy as a function of NFE under different supervision update strategies: one-shot self-evolution (100\%), progressive self-evolution with mixing ratios of 10\% and 1\%, and progressive self-evolution (1\%) without trust-region regularization.
Each curve shows the resulting accuracy-efficiency frontier under a fixed supervision evolution scheme.
}
\label{fig:progressive_ablation}
\end{center}
\end{figure}

Refinement regulation reshapes the decoding process itself.
As the controller changes how tokens converge, it simultaneously alters the refinement trajectories that define supervision for subsequent stages, inducing a self-generated supervision shift.
We therefore ablate both how supervision is updated across stages and whether trust-region regularization is applied.

Figure~\ref{fig:progressive_ablation} compares one-shot self-evolution, which fully replaces supervision trajectories at each stage, with progressive self-evolution that gradually incorporates rollouts from the current regulated decoder.
Progressive self-evolution consistently achieves stronger accuracy-efficiency trade-offs across decoding budgets.
In particular, a small mixing ratio (1\%) performs best, outperforming both direct replacement (100\%) and larger ratios (10\%).
This indicates that supervision must track the evolving refinement process, but abrupt distributional shifts make controller learning less reliable, whereas gradual trajectory evolution provides more reliable training signals.

Removing trust-region regularization significantly degrades performance.
Without constraining successive controllers, refinement behavior changes too abruptly, leading to inferior accuracy at comparable NFE.
Together, these results show that effective refinement regulation requires both progressive self-evolving supervision to follow induced trajectory shifts and trust-region regularization to control the rate of process change.

\subsection{Effect of Refinement Regulation Strength}
\label{sec:alpha_ablation}

\begin{figure}[t]
\begin{center}
\includegraphics[width=\columnwidth]{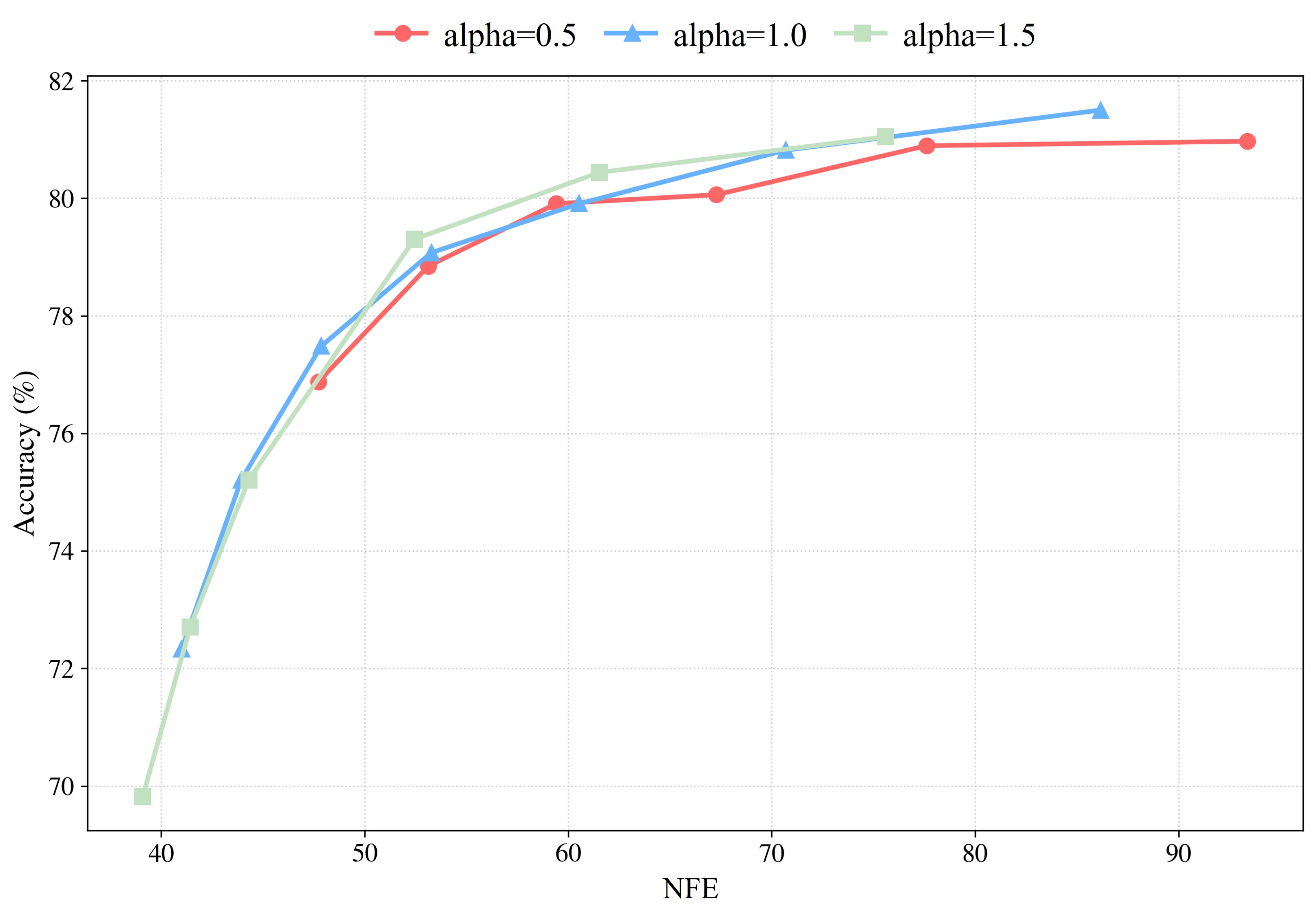}
\caption{
Effect of refinement regulation strength $\alpha$ on the quality-efficiency trade-off.
We report accuracy as a function of the number of function evaluations (NFE) under $\alpha \in \{0.5, 1.0, 1.5\}$.
Larger $\alpha$ enforces stronger temperature-based refinement regulation, leading to earlier token unmasking and lower NFE, while smaller $\alpha$ yields more conservative refinement.
Each curve traces the resulting accuracy-NFE frontier induced by a fixed $\alpha$.
}
\label{fig:alpha_ablation}
\end{center}
\end{figure}

PRR converts predicted empirical convergence progress into refinement suppression through a regulation function parameterized by $\alpha$.
Larger $\alpha$ induces stronger temperature modulation, sharpening the predictive distribution for tokens with high predicted refinement progress and thereby promoting earlier unmasking, while smaller $\alpha$ results in more conservative refinement.

Figure~\ref{fig:alpha_ablation} shows the resulting quality-efficiency frontiers.
Increasing $\alpha$ consistently shifts the frontier toward lower NFE, yielding stronger acceleration in the low-budget regime.
Conversely, smaller $\alpha$ slightly favors final accuracy when more refinement steps are allowed.
Importantly, all settings produce smooth and well-ordered frontiers, with no abrupt performance drops or instability.

These results indicate that PRR does not rely on a narrowly tuned operating point.
Instead, $\alpha$ serves as a continuous refinement control knob, allowing practitioners to flexibly trade off decoding efficiency and generation quality without retraining the diffusion model.

\section{Demonstration of Refinement Dynamics of PRR vs.\ LLaDA}
\label{demo}

We visualize and analyze the refinement dynamics induced by PRR (built on LLaDA) in comparison with the standard LLaDA baseline on four mathematical reasoning examples from GSM8K.
The LLaDA baseline adopts the standard top-1 unmasking rule.
For clarity, we do not visualize refinement steps for mathematical expression tokens.
This is because mathematical expressions contain many non-semantic or formatting-related tokens (e.g., line breaks, separators, and symbol-level fragments) that are not meaningful to interpret at the word level and would significantly clutter the visualization.
For example, even a simple expression such as ``$1/2$'' may correspond to multiple low-level tokens and require several refinement steps.
We therefore restrict the visualization to natural language tokens when annotating refinement dynamics.

\begin{questionbox}{Question 1} \\
Janet's ducks lay 16 eggs per day. She eats three for breakfast every morning and bakes muffins for her friends every day with four. She sells the remainder at the farmers' market daily for \$2 per fresh duck egg. How much in dollars does she make every day at the farmers' market?
\end{questionbox}

\begin{answerbox}\\
Janet sells $16 - 3 - 4 = 9$ duck eggs a day.\\
She makes $9 \times 2 = 18$ dollars every day at the farmer's market.

\textbf{Answer:} 18
\end{answerbox}

\begin{baselinebox}{Baseline reasoning process}\\
\stepword{To}{2} \stepword{determine}{1} \stepword{how}{3} \stepword{much}{4} \stepword{Janet}{5} \stepword{makes}{6} \stepword{every}{7} \stepword{day}{8} \stepword{at}{14} \stepword{the}{9} \stepword{farmers}{13}\stepword{'}{11} \stepword{market}{10}\stepword{,}{12} \stepword{we}{15} \stepword{need}{17} \stepword{to}{16} \stepword{follow}{20} \stepword{these}{22} \stepword{steps}{21}\stepword{:}{19}

\par
\stepword{Step}{87} \stepword{1}{86}\stepword{:}{86} \stepword{Janet}{88}\stepword{'}{96} \stepword{ducks}{90} \stepword{lay}{89} 16 \stepword{eggs}{91} \stepword{per}{95} \stepword{day}{97}\stepword{.}{98}

\par
\stepword{Step}{100} \stepword{2}{101}\stepword{:}{101} \stepword{She}{128} \stepword{eats}{102} 3 \stepword{eggs}{105} \stepword{for}{107} \stepword{breakfast}{106} \stepword{every}{109} \stepword{morning}{108}\stepword{.}{111}

\par
\stepword{Step}{126} \stepword{3}{125}\stepword{:}{125} \stepword{She}{117} \stepword{bakes}{112} 4 \stepword{eggs}{116} \stepword{for}{118} \stepword{her}{120} \stepword{friends}{119} \stepword{every}{121} \stepword{day}{123}\stepword{.}{122}

\par
\stepword{So}{140}\stepword{,}{129} \stepword{the}{138} \stepword{total}{139} \stepword{number}{136} \stepword{of}{135} \stepword{eggs}{134} \stepword{she}{137} \stepword{eats}{133} \stepword{and}{132} \stepword{bakes}{130} \stepword{is}{143}\stepword{:}{142}
\[
3 + 4 = 7
\]

\par
\stepword{Step}{158} \stepword{4}{157}\stepword{:}{157} \stepword{The}{159} \stepword{number}{160} \stepword{of}{161} \stepword{eggs}{162} \stepword{left}{163} \stepword{to}{192} \stepword{sell}{191} \stepword{is}{180}\stepword{:}{173}
\[
16 - 7 = 9
\]

\par
\stepword{Step}{181} \stepword{5}{182}\stepword{:}{182} \stepword{Janet}{184} \stepword{sells}{183} \stepword{each}{185} \stepword{egg}{187} \stepword{for}{189} \stepword{\$2}{189}\stepword{.}{190} \stepword{Therefore}{193}\stepword{,}{194} \stepword{the}{198} \stepword{revenue}{195} \stepword{from}{196} \stepword{selling}{197} 9 \stepword{eggs}{201} \stepword{is}{204}\stepword{:}{203}
\[
9 \times 2 = 18
\]

\par
\stepword{So}{223}\stepword{,}{221} \stepword{Janet}{220} \stepword{makes}{222} \stepword{\$18}{222} \stepword{every}{229} \stepword{day}{228} \stepword{at}{230} \stepword{the}{231} \stepword{farmers}{235}\stepword{'}{233} \stepword{market}{232}\stepword{.}{234} \finalanswer{18}
\end{baselinebox}

\begin{metricsbox} \\
\textbullet\ Latency: 48.4855s\\
\textbullet\ NFE (Steps): 256
\end{metricsbox}

\begin{reasoningbox}{PRR reasoning process}\\
\stepword{To}{2} \stepword{determine}{1} \stepword{how}{2} \stepword{much}{2} \stepword{Janet}{1} \stepword{makes}{2} \stepword{every}{2} \stepword{day}{2} \stepword{at}{2} \stepword{the}{2} \stepword{farmers}{2}\stepword{'}{2} \stepword{market}{2}\stepword{,}{2} \stepword{we}{2} \stepword{need}{2} \stepword{to}{2} \stepword{follow}{3} \stepword{these}{3} \stepword{steps}{3}\stepword{:}{2}

\par
\stepword{Step}{29} \stepword{1}{28}\stepword{:}{28} \stepword{The}{31} \stepword{total}{31} \stepword{number}{29} \stepword{of}{30} \stepword{eggs}{30} \stepword{laid}{30} \stepword{by}{33} \stepword{Janet}{33}\stepword{'}{33} \stepword{ducks}{33} \stepword{per}{33} \stepword{day}{33} \stepword{is}{33} 16\stepword{.}{35}

\par
\stepword{Step}{36} \stepword{2}{36}\stepword{:}{36} \stepword{Janet}{36} \stepword{eats}{36} 3 \stepword{eggs}{36} \stepword{for}{37} \stepword{breakfast}{37} \stepword{every}{39} \stepword{morning}{38}\stepword{.}{38} \stepword{She}{38} \stepword{bakes}{38} 4 \stepword{eggs}{40} \stepword{for}{41} \stepword{her}{41} \stepword{friends}{42} \stepword{every}{43} \stepword{day}{42}\stepword{.}{45} \stepword{Therefore}{46}\stepword{,}{45} \stepword{she}{46} \stepword{eats}{47} \stepword{a}{48} \stepword{total}{47} \stepword{of}{48} $(3 + 4 = 7)$ \stepword{eggs}{48} \stepword{per}{50} \stepword{day}{51}\stepword{.}{50}

\par
\stepword{Step}{52} \stepword{3}{52}\stepword{:}{52} \stepword{The}{53} \stepword{number}{54} \stepword{of}{54} \stepword{eggs}{53} \stepword{left}{55} \stepword{for}{56} \stepword{the}{55} \stepword{farmers}{56}\stepword{'}{58} \stepword{market}{58} \stepword{is}{58} $(16 - 7 = 9)$ \stepword{eggs}{61} \stepword{per}{61} \stepword{day}{62}\stepword{.}{62}

\par
\stepword{Step}{63} \stepword{4}{64}\stepword{:}{64} \stepword{Janet}{65} \stepword{sells}{64} \stepword{each}{65} \stepword{egg}{67} \stepword{for}{66} \stepword{\$2}{66}\stepword{.}{67} \stepword{Therefore}{68}\stepword{,}{68} \stepword{her}{67} \stepword{daily}{68} \stepword{earnings}{70} \stepword{from}{71} \stepword{selling}{70} \stepword{the}{72} \stepword{eggs}{72} \stepword{are}{74} $(9 \times 2 = 18)$ \stepword{dollars}{71}\stepword{.}{72}

\par
\stepword{So}{74}\stepword{,}{72} \stepword{Janet}{73} \stepword{makes}{73} \stepword{\$18}{73} \stepword{every}{76} \stepword{day}{76} \stepword{at}{76} \stepword{the}{76} \stepword{farmers}{76}\stepword{'}{76} \stepword{market}{76}\stepword{.}{76} \finalanswer{18}
\end{reasoningbox}

\begin{metricsbox} \\
\textbullet\ Latency: 14.1445s\\
\textbullet\ NFE (Steps): 77\\
\textbullet\ Speedup: 3.43x (Latency)\\
\textbullet\ Speedup: 3.32x (NFE)
\end{metricsbox}

\vspace{1.5em}

\begin{questionbox}{Question 2} \\
A robe takes 2 bolts of blue fiber and half that much white fiber. How many bolts in total does it take?
\end{questionbox}

\begin{answerbox}\\
White fiber required: $\frac{2}{2} = 1$ bolt.\\
Total fabric: $2 + 1 = 3$ bolts.

\textbf{Answer:} 3
\end{answerbox}

\begin{baselinebox}{Baseline reasoning process}\\
\stepword{To}{3} \stepword{determine}{2} \stepword{the}{4} \stepword{total}{5} \stepword{number}{7} \stepword{of}{1} \stepword{bolts}{6} \stepword{required}{30} \stepword{for}{14} \stepword{the}{10} \stepword{robe}{11}\stepword{,}{8} \stepword{we}{9} \stepword{need}{13} \stepword{to}{12} \stepword{add}{25} \stepword{the}{17} \stepword{amount}{19} \stepword{of}{16} \stepword{blue}{15} \stepword{fiber}{18} \stepword{to}{29} \stepword{the}{21} \stepword{amount}{20} \stepword{of}{22} \stepword{white}{23} \stepword{fiber}{24}\stepword{.}{26}

\par
\stepword{Given}{32}\stepword{:}{31}
\stepword{-}{34} \stepword{The}{45} \stepword{robe}{38} \stepword{takes}{47} 2 \stepword{bolts}{37} \stepword{of}{39} \stepword{blue}{40} \stepword{fiber}{43}\stepword{.}{42}
\stepword{-}{44} \stepword{The}{46} \stepword{robe}{49} \stepword{takes}{48} \stepword{half}{50} \stepword{as}{55} \stepword{much}{51} \stepword{white}{53} \stepword{fiber}{52} \stepword{as}{54} \stepword{blue}{56} \stepword{fiber}{57}\stepword{.}{58}

\par
\stepword{First}{61}\stepword{,}{62} \stepword{let}{64}\stepword{'}{63} \stepword{s}{63} \stepword{calculate}{70} \stepword{the}{66} \stepword{amount}{67} \stepword{of}{65} \stepword{white}{69} \stepword{fiber}{68}\stepword{:}{71}
\[
\text{White fiber} = \frac{1}{2} \times \text{Blue fiber}
\]
\[
\text{White fiber} = \frac{1}{2} \times 2 = 1 \text{ bolt}
\]

\par
\stepword{Next}{147}\stepword{,}{140} \stepword{we}{142} \stepword{add}{141} \stepword{the}{143} \stepword{amount}{153} \stepword{of}{144} \stepword{blue}{145} \stepword{fiber}{146} \stepword{to}{160} \stepword{the}{149} \stepword{amount}{148} \stepword{of}{150} \stepword{white}{151} \stepword{fiber}{152} \stepword{to}{154} \stepword{find}{157} \stepword{the}{155} \stepword{total}{156} \stepword{number}{159} \stepword{of}{158} \stepword{bolts}{161}\stepword{:}{162}
\[
\text{Total bolts} = \text{Blue fiber} + \text{White fiber} = 2 + 1 = 3 \text{ bolts}
\]

\par
\stepword{Therefore}{224}\stepword{,}{222} \stepword{the}{237} \stepword{robe}{249} \stepword{takes}{252} \stepword{a}{246} \stepword{total}{245} \stepword{of}{244} 3 \stepword{bolts}{248} \stepword{of}{255} \stepword{fiber}{256}\stepword{.}{254} \finalanswer{3}
\end{baselinebox}

\begin{metricsbox} \\
\textbullet\ Latency: 46.5072s\\
\textbullet\ NFE (Steps): 256
\end{metricsbox}

\begin{reasoningbox}{PRR reasoning process}\\
\stepword{To}{2} \stepword{determine}{1} \stepword{the}{2} \stepword{total}{2} \stepword{number}{2} \stepword{of}{1} \stepword{bolts}{2} \stepword{required}{5} \stepword{for}{4} \stepword{the}{4} \stepword{robe}{4}\stepword{,}{3} \stepword{we}{3} \stepword{need}{4} \stepword{to}{4} \stepword{add}{9} \stepword{the}{6} \stepword{amount}{7} \stepword{of}{6} \stepword{blue}{5} \stepword{fiber}{7} \stepword{to}{11} \stepword{the}{8} \stepword{amount}{8} \stepword{of}{9} \stepword{white}{9} \stepword{fiber}{9}\stepword{.}{10}

\par
\stepword{Step}{12} \stepword{1}{12}\stepword{:}{12} \stepword{The}{14} \stepword{robe}{15} \stepword{takes}{15} 2 \stepword{bolts}{15} \stepword{of}{15} \stepword{blue}{15} \stepword{fiber}{15}\stepword{.}{15}

\par
\stepword{Step}{15} \stepword{2}{15}\stepword{:}{15} \stepword{The}{15} \stepword{robe}{16} \stepword{takes}{16} \stepword{half}{17} \stepword{as}{18} \stepword{much}{16} \stepword{white}{17} \stepword{fiber}{16} \stepword{as}{17} \stepword{blue}{18} \stepword{fiber}{19}\stepword{.}{18} \stepword{Since}{19} \stepword{it}{21} \stepword{takes}{21} 2 \stepword{bolts}{21} \stepword{of}{21} \stepword{blue}{23} \stepword{fiber}{23}\stepword{,}{23} \stepword{half}{24} \stepword{of}{24} \stepword{that}{24} \stepword{is}{25} $\frac{2}{2} = 1$ \stepword{bolt}{27} \stepword{of}{27} \stepword{white}{26} \stepword{fiber}{27}\stepword{.}{27}

\par
\stepword{Now}{27}\stepword{,}{27} \stepword{we}{29} \stepword{add}{28} \stepword{the}{28} \stepword{bolts}{30} \stepword{of}{29} \stepword{blue}{30} \stepword{fiber}{32} \stepword{and}{32} \stepword{white}{33} \stepword{fiber}{33}\stepword{:}{35}
\[
2 \text{ bolts of blue fiber} + 1 \text{ bolt of white fiber} = 3 \text{ bolts}
\]

\par
\stepword{Therefore}{46}\stepword{,}{42} \stepword{the}{43} \stepword{robe}{44} \stepword{takes}{44} \stepword{a}{44} \stepword{total}{44} \stepword{of}{43} 3 \stepword{bolts}{44}\stepword{.}{44} \finalanswer{3}
\end{reasoningbox}

\begin{metricsbox} \\
\textbullet\ Latency: 9.6268s\\
\textbullet\ NFE (Steps): 53\\
\textbullet\ Speedup: 4.83x (Latency)\\
\textbullet\ Speedup: 4.83x (NFE)
\end{metricsbox}

\vspace{1.5em}

\begin{questionbox}{Question 3} \\
James decides to run 3 sprints 3 times a week. He runs 60 meters each sprint. How many total meters does he run a week?
\end{questionbox}

\begin{answerbox}\\
Total sprints per week: $3 \times 3 = 9$\\
Total distance: $9 \times 60 = 540$ meters.

\textbf{Answer:} 540
\end{answerbox}

\begin{baselinebox}{Baseline reasoning process}\\
\stepword{To}{2} \stepword{determine}{3} \stepword{the}{4} \stepword{total}{1} \stepword{distance}{12} \stepword{James}{6} \stepword{runs}{5} \stepword{in}{8} \stepword{a}{10} \stepword{week}{7}\stepword{,}{9} \stepword{we}{11} \stepword{need}{14} \stepword{to}{13} \stepword{break}{16} \stepword{down}{15} \stepword{the}{17} \stepword{problem}{18} \stepword{into}{20} \stepword{smaller}{19} \stepword{steps}{21}\stepword{.}{24}

\par
\stepword{Step}{30} \stepword{1}{31}\stepword{:}{29} \stepword{Calculate}{32} \stepword{the}{27} \stepword{distance}{25} \stepword{James}{26} \stepword{runs}{33} \stepword{in}{34} \stepword{one}{35} \stepword{day}{54}\stepword{.}{37}\\
\stepword{James}{38} \stepword{runs}{39} 3 \stepword{sprints}{42} \stepword{each}{44} \stepword{day}{55}\stepword{,}{52} \stepword{and}{53} \stepword{each}{47} \stepword{sprint}{45} \stepword{is}{46} 60 \stepword{meters}{50} \stepword{long}{56}\stepword{.}{57}
\begin{align*}
\text{Distance per day} &= 3 \text{ sprints} \times 60 \text{ meters/sprint} \\
&= 180 \text{ meters/day}
\end{align*}

\par
\stepword{Step}{100} \stepword{2}{105}\stepword{:}{103} \stepword{Calculate}{112} \stepword{the}{110} \stepword{distance}{111} \stepword{James}{99} \stepword{runs}{98} \stepword{in}{109} \stepword{one}{115} \stepword{week}{116}\stepword{.}{113}\\
\stepword{James}{120} \stepword{runs}{119} 3 \stepword{times}{128} \stepword{a}{122} \stepword{week}{121}\stepword{.}{125}
\begin{align*}
\text{Distance per week} &= 180 \text{ meters/day} \times 3 \text{ days/week} \\
&= 540 \text{ meters/week}
\end{align*}

\par
\stepword{Therefore}{186}\stepword{,}{170} \stepword{the}{172} \stepword{total}{173} \stepword{distance}{175} \stepword{James}{171} \stepword{runs}{174} \stepword{in}{178} \stepword{a}{179} \stepword{week}{177} \stepword{is}{176} 540 \stepword{meters}{188}\stepword{.}{190} \finalanswer{540}
\end{baselinebox}

\begin{metricsbox} \\
\textbullet\ Latency: 47.0138s\\
\textbullet\ NFE (Steps): 256
\end{metricsbox}

\begin{reasoningbox}{PRR reasoning process}\\
\stepword{To}{2} \stepword{determine}{1} \stepword{the}{2} \stepword{total}{1} \stepword{number}{5} \stepword{of}{4} \stepword{runs}{3} \stepword{James}{5} \stepword{runs}{12} \stepword{in}{4} \stepword{a}{4} \stepword{week}{3}\stepword{,}{4} \stepword{we}{4} \stepword{need}{7} \stepword{to}{6} \stepword{follow}{7} \stepword{these}{7} \stepword{steps}{7}\stepword{:}{7}

\par
\stepword{Step}{32} \stepword{1}{32}\stepword{:}{32} \stepword{Calculate}{32} \stepword{the}{30} \stepword{number}{30} \stepword{of}{30} \stepword{sprints}{30} \stepword{James}{30} \stepword{runs}{30} \stepword{in}{30} \stepword{a}{30} \stepword{week}{30}\stepword{.}{32}\\
\stepword{James}{32} \stepword{runs}{32} 3 \stepword{sprints}{32} \stepword{each}{33} \stepword{day}{33} \stepword{and}{34} \stepword{he}{36} \stepword{does}{36} \stepword{this}{36} 3 \stepword{times}{39} \stepword{in}{38} \stepword{a}{38} \stepword{week}{37}\stepword{.}{38}
\[
3 \text{ sprints/day} \times 3 \text{ days/week} = 9 \text{ sprints/week}
\]

\par
\stepword{Step}{44} \stepword{2}{44}\stepword{:}{44} \stepword{Determine}{44} \stepword{the}{44} \stepword{total}{44} \stepword{distance}{45} \stepword{James}{44} \stepword{runs}{44} \stepword{in}{45} \stepword{one}{44} \stepword{sprint}{44}\stepword{.}{44}\\
\stepword{James}{45} \stepword{runs}{45} 60 \stepword{meters}{48} \stepword{each}{50} \stepword{sprint}{50}\stepword{.}{50}

\par
\stepword{Step}{50} \stepword{3}{50}\stepword{:}{50} \stepword{Multiply}{50} \stepword{the}{50} \stepword{number}{50} \stepword{of}{50} \stepword{sprints}{50} \stepword{in}{50} \stepword{a}{50} \stepword{week}{50} \stepword{by}{50} \stepword{the}{50} \stepword{distance}{50} \stepword{of}{50} \stepword{one}{50} \stepword{sprint}{50}\stepword{.}{49}
\[
9 \text{ sprints/week} \times 60 \text{ meters/sprint} = 540 \text{ meters/week}
\]

\par
\stepword{Therefore}{56}\stepword{,}{56} \stepword{the}{56} \stepword{total}{56} \stepword{number}{57} \stepword{of}{57} \stepword{meters}{57} \stepword{James}{56} \stepword{runs}{57} \stepword{in}{57} \stepword{a}{57} \stepword{week}{57} \stepword{is}{57} 540\finalanswer{540}
\end{reasoningbox}

\begin{metricsbox} \\
\textbullet\ Latency: 10.9641s\\
\textbullet\ NFE (Steps): 60\\
\textbullet\ Speedup: 4.29x (Latency)\\
\textbullet\ Speedup: 4.27x (NFE)
\end{metricsbox}

\vspace{1.5em}

\begin{questionbox}{Question 4} \\
Every day, Wendi feeds each of her chickens three cups of mixed chicken feed, containing seeds, mealworms and vegetables to help keep them healthy. She gives the chickens their feed in three separate meals. In the morning, she gives her flock of chickens 15 cups of feed. In the afternoon, she gives her chickens another 25 cups of feed. How many cups of feed does she need to give her chickens in the final meal of the day if the size of Wendi's flock is 20 chickens?
\end{questionbox}

\begin{answerbox}\\
Total feed needed per day: $3 \times 20 = 60$ cups.\\
Feed given in morning and afternoon: $15 + 25 = 40$ cups.\\
Feed needed for final meal: $60 - 40 = 20$ cups.

\textbf{Answer:} 20
\end{answerbox}

\begin{baselinebox}{Baseline reasoning process}\\
\stepword{To}{2} \stepword{determine}{1} \stepword{how}{8} \stepword{many}{6} \stepword{cups}{9} \stepword{of}{7} \stepword{feed}{5} \stepword{Wendi}{3} \stepword{needs}{10} \stepword{to}{11} \stepword{give}{12} \stepword{her}{14} \stepword{chickens}{13} \stepword{in}{18} \stepword{the}{17} \stepword{final}{16} \stepword{meal}{15} \stepword{of}{23} \stepword{the}{20} \stepword{day}{21}\stepword{,}{22} \stepword{we}{19} \stepword{can}{30} \stepword{follow}{25} \stepword{these}{29} \stepword{steps}{26}\stepword{:}{27}

\par
\stepword{Step}{69} \stepword{1}{70}\stepword{:}{70} \stepword{Calculate}{83} \stepword{the}{75} \stepword{total}{74} \stepword{amount}{73} \stepword{of}{76} \stepword{feed}{77} \stepword{needed}{78} \stepword{for}{79} \stepword{the}{80} \stepword{entire}{81} \stepword{day}{82}\stepword{:}{91}\\
\stepword{-}{93} \stepword{Each}{92} \stepword{chicken}{84} \stepword{needs}{96} 3 \stepword{cups}{86} \stepword{of}{90} \stepword{feed}{89} \stepword{per}{94} \stepword{day}{95}\stepword{.}{97}\\
\stepword{-}{99} \stepword{Wendi}{101} \stepword{has}{106} 20 \stepword{chickens}{102}\stepword{.}{108}\\
\stepword{-}{112} \stepword{Total}{111} \stepword{feed}{110} \stepword{needed}{109} $= 20 \text{ chickens} \times 3 \text{ cups/chicken} = 60 \text{ cups}$

\par
\stepword{Step}{133} \stepword{2}{130}\stepword{:}{130} \stepword{Subtract}{142} \stepword{the}{140} \stepword{amount}{141} \stepword{of}{138} \stepword{feed}{139} \stepword{already}{136} \stepword{given}{137} \stepword{in}{145} \stepword{the}{144} \stepword{morning}{146} \stepword{and}{143} \stepword{afternoon}{147} \stepword{from}{149} \stepword{the}{150} \stepword{total}{151} \stepword{amount}{152} \stepword{needed}{148}\stepword{:}{153}\\
\stepword{-}{155} \stepword{Feed}{160} \stepword{given}{157} \stepword{in}{156} \stepword{the}{159} \stepword{morning}{158}\stepword{:}{192} 15 \stepword{cups}{161}\\
\stepword{-}{174} \stepword{Feed}{176} \stepword{given}{172} \stepword{in}{167} \stepword{the}{166} \stepword{afternoon}{165}\stepword{:}{191} 25 \stepword{cups}{168}\\
\stepword{-}{178} \stepword{Total}{177} \stepword{feed}{179} \stepword{given}{180} \stepword{in}{188} \stepword{morning}{185} \stepword{and}{187} \stepword{afternoon}{184} $= 15 \text{ cups} + 25 \text{ cups} = 40 \text{ cups}$

\par
\stepword{Step}{205} \stepword{3}{204}\stepword{:}{204} \stepword{Subtract}{206} \stepword{this}{207} \stepword{from}{211} \stepword{the}{210} \stepword{total}{209} \stepword{feed}{215} \stepword{needed}{208}\stepword{:}{213}\\
\stepword{-}{214} \stepword{Feed}{217} \stepword{needed}{216} \stepword{for}{224} \stepword{the}{219} \stepword{final}{218} \stepword{meal}{220} $= 60 \text{ cups} - 40 \text{ cups} = 20 \text{ cups}$

\par
\stepword{Therefore}{240}\stepword{,}{242} \stepword{Wendi}{244} \stepword{needs}{245} \stepword{to}{246} \stepword{give}{249} \stepword{her}{253} \stepword{chickens}{255} 20 \stepword{cups}{251} \stepword{of}{248} \stepword{feed}{247}\stepword{.}{256} \finalanswer{20}
\end{baselinebox}

\begin{metricsbox} \\
\textbullet\ Latency: 49.1492s\\
\textbullet\ NFE (Steps): 256
\end{metricsbox}

\begin{reasoningbox}{PRR reasoning process}\\
\stepword{To}{2} \stepword{determine}{1} \stepword{how}{1} \stepword{many}{2} \stepword{cups}{2} \stepword{of}{2} \stepword{feed}{2} \stepword{Wendi}{2} \stepword{needs}{2} \stepword{to}{2} \stepword{give}{2} \stepword{her}{2} \stepword{chickens}{2} \stepword{in}{2} \stepword{the}{2} \stepword{final}{2} \stepword{meal}{2} \stepword{of}{3} \stepword{the}{4} \stepword{day}{4}\stepword{,}{4} \stepword{we}{3} \stepword{can}{6} \stepword{follow}{6} \stepword{these}{6} \stepword{steps}{5}\stepword{:}{6}

\par
\stepword{Step}{22} \stepword{1}{22}\stepword{:}{22} \stepword{Calculate}{24} \stepword{the}{24} \stepword{total}{24} \stepword{amount}{24} \stepword{of}{24} \stepword{feed}{24} \stepword{needed}{23} \stepword{for}{24} \stepword{the}{24} \stepword{day}{24}\stepword{:}{25}\\
\stepword{-}{26} \stepword{Each}{26} \stepword{chicken}{26} \stepword{gets}{29} 3 \stepword{cups}{25} \stepword{of}{26} \stepword{feed}{26} \stepword{per}{27} \stepword{day}{28}\stepword{.}{28}\\
\stepword{-}{28} \stepword{Wendi}{31} \stepword{has}{32} 20 \stepword{chickens}{31}\stepword{.}{33}\\
\stepword{-}{35} \stepword{Total}{34} \stepword{feed}{34} \stepword{needed}{33} $= 20 \text{ chickens} \times 3 \text{ cups/chicken} = 60 \text{ cups}$

\par
\stepword{Step}{42} \stepword{2}{42}\stepword{:}{42} \stepword{Subtract}{45} \stepword{the}{44} \stepword{amount}{45} \stepword{of}{45} \stepword{feed}{45} \stepword{already}{44} \stepword{given}{45} \stepword{in}{46} \stepword{the}{46} \stepword{morning}{46} \stepword{and}{45} \stepword{afternoon}{46} \stepword{from}{48} \stepword{the}{48} \stepword{total}{47} \stepword{amount}{48} \stepword{needed}{47}\stepword{:}{48}\\
\stepword{-}{48} \stepword{Feed}{50} \stepword{given}{50} \stepword{in}{49} \stepword{the}{50} \stepword{morning}{49} $= 15 \text{ cups}$\\
\stepword{-}{53} \stepword{Feed}{59} \stepword{given}{56} \stepword{in}{56} \stepword{the}{56} \stepword{afternoon}{53} $= 25 \text{ cups}$\\
\stepword{-}{55} \stepword{Total}{54} \stepword{feed}{55} \stepword{given}{55} \stepword{so}{58} \stepword{far}{58} $= 15 \text{ cups} + 25 \text{ cups} = 40 \text{ cups}$

\par
\stepword{Step}{63} \stepword{3}{62}\stepword{:}{62} \stepword{The}{63} \stepword{amount}{64} \stepword{needed}{65} \stepword{for}{65} \stepword{the}{65} \stepword{final}{65} \stepword{meal}{64} \stepword{is}{65}\stepword{:}{65}\\
\stepword{-}{66} $60 \text{ cups} - 40 \text{ cups} = 20 \text{ cups}$

\par
\stepword{Therefore}{70}\stepword{,}{68} \stepword{Wendi}{68} \stepword{needs}{68} \stepword{to}{68} \stepword{give}{68} \stepword{her}{69} \stepword{chickens}{69} 20 \stepword{cups}{69} \stepword{of}{69} \stepword{feed}{69} \stepword{in}{68} \stepword{the}{68} \stepword{final}{68} \stepword{meal}{68} \stepword{of}{68} \stepword{the}{68} \stepword{day}{68}\stepword{.}{68} \finalanswer{20}
\end{reasoningbox}

\begin{metricsbox} \\
\textbullet\ Latency: 13.5965s\\
\textbullet\ NFE (Steps): 71\\
\textbullet\ Speedup: 3.61x (Latency)\\
\textbullet\ Speedup: 3.61x (NFE)
\end{metricsbox}

\end{document}